\documentclass[preprint,12pt]{elsarticle}
\journal{Mechanical Systems and Signal Processing}

\newcommand{\R}{\mathbb{R}}

\usepackage[hidelinks]{hyperref}
\usepackage{amsmath,amsfonts}
\usepackage{amssymb}
\usepackage{algorithmic}
\usepackage{array}
\usepackage{enumitem}
\usepackage{textcomp}
\usepackage{stfloats}
\usepackage{subcaption} 
\usepackage{url}
\usepackage{verbatim}
\usepackage{graphicx}
\usepackage{tabularx}
\usepackage{booktabs}
\usepackage{colortbl}
\usepackage{multirow}
\usepackage{siunitx}
\usepackage{cleveref} 
\def\BibTeX{{\rm B\kern-.05em{\sc i\kern-.025em b}\kern-.08em
    T\kern-.1667em\lower.7ex\hbox{E}\kern-.125emX}}
\usepackage{balance}
\usepackage{xcolor}
\hypersetup{
    colorlinks=true,
    linkcolor=cyan, 
    filecolor=cyan, 
    urlcolor=cyan, 
    citecolor=orange
}
\newcommand{\revtext}[1]{{\color{black}#1}}

\begin{document}

\begin{frontmatter}


\title{Graph Neural Networks for Virtual Sensing \\in Complex Systems: \\Addressing Heterogeneous Temporal Dynamics}

\author[inst1]{Mengjie Zhao\corref{cor1}}
\ead{mengjie.zhao@epfl.ch}

\affiliation[inst1]{organization={Intelligent Maintenance and Operations Systems, EPFL},
            city={Lausanne},
            country={Switzerland}}

\author[inst2]{Cees Taal}
\ead{cees.taal@skf.com}
\author[inst2]{Stephan Baggerohr}
\ead{stephan.baggerohr@skf.com}
\author[inst1]{Olga Fink\corref{cor1}}
\ead{olga.fink@epfl.ch}

\affiliation[inst2]{organization={SKF, Research and Technology Development},
            city={Houten},
            country={Netherlands}}

\cortext[cor1]{Corresponding authors.}

\begin{abstract}%
Real-time condition monitoring is crucial for the reliable and efficient operation of complex systems.
However, relying solely on physical sensors can be limited due to their cost, placement constraints, or inability to directly measure certain critical parameters. Virtual sensing addresses these limitations by leveraging readily available sensor data and system knowledge to estimate inaccessible parameters or infer system states.
The increasing complexity of industrial systems necessitates deployments of sensors with diverse modalities to provide a comprehensive understanding of system states. These sensors capture data at varying frequencies to monitor both rapid and slowly varying system dynamics, as well as local and global state evolutions of the systems. However, this leads to heterogeneous temporal dynamics, which, particularly under varying operating conditions, pose a significant challenge for accurate virtual sensing.
To address these challenges, we propose a novel Heterogeneous Temporal Graph Neural Network (HTGNN) framework for virtual sensing. HTGNN explicitly models signals from diverse sensors as distinct node types within a graph structure, enabling the capture of complex relationships between sensors. Additionally, HTGNN integrates context from operating conditions, derived from exogenous variables such as control settings and external environmental factors into the model architecture. This integration allows HTGNN to adapt to diverse operating and environmental conditions, ensuring accurate and robust virtual sensing.
We evaluate the effectiveness of HTGNN using two newly released, publicly available datasets: a test-rig bearing dataset with diverse load conditions for bearing load prediction and a comprehensive year-long simulated dataset for train-bridge-track interaction, aimed at predicting bridge live loads. Our extensive experiments demonstrate that HTGNN significantly outperforms established baseline methods in both bearing and bridge load prediction tasks, particularly under highly varying operating conditions. These results highlight HTGNN's potential as a robust and accurate virtual sensing approach for complex systems, paving the way for improved monitoring, predictive maintenance, and enhanced system performance. Our code and data are available under \url{https://github.com/EPFL-IMOS/htgnn}.

\end{abstract}

\begin{highlights}
\item Introduces novel Heterogeneous Temporal Graph Neural Network for virtual sensing in complex systems with diverse sensor modalities.

\item Effectively integrates information from sensors with both intra-modality and inter-modality interactions.

\item Captures the influence of exogenous variables on sensor signals for robust prediction.

\item Releases bearing and train-track-bridge datasets for virtual sensing research.

\end{highlights}



\end{frontmatter}

\section{Introduction}
Real-time condition monitoring is crucial for ensuring the smooth, efficient, and safe operation of critical industrial and infrastructure systems across diverse sectors~\cite{soleimani2021diagnostics}. The information gathered from monitoring system states can be used to optimize operational performance, proactively detect faults to prevent costly failures, and extend the lifetime of industrial and infrastructure assets~\cite{fink2020potential}. 
However, conventional physical sensors have several limitations, including power supply, connectivity, durability and reliability, as well as coverage and resolution~\cite{yoon2022virtual}.
Some critical parameters, such as internal stresses within components, are often inaccessible without intrusive or destructive testing. Harsh environmental conditions, including extreme temperatures, vibrations, or corrosive chemicals, can significantly reduce sensor accuracy and lifespan, introducing noise and drift into the measurements~\cite{teh2020sensor}. 
Additionally, although application-specific sensors have been developed to meet specific monitoring needs, their high cost often restricts their widespread deployment. Furthermore, achieving comprehensive coverage and scalability with physical sensors is challenging due to the expense and logistical complexities associated with installing sensors in all required locations, which can be prohibitive. This can result in potential blind spots in the monitoring system, undermining its effectiveness. 

To overcome these challenges, \textbf{virtual sensing} has emerged as a promising alternative. By leveraging readily available sensor data and system knowledge, \textit{virtual sensing} can estimate parameters that are otherwise inaccessible or compensate for limitations in existing sensors. 
The capability to accurately estimate critical variables is essential for enabling more informed decision-making and proactive maintenance strategies across various applications.
For instance, in rotating machinery, accurate virtual sensing of bearing load during operation is a key enabler for optimizing performance, detecting misalignments~\cite{widner1976bearing}, diagnosing faults, and predicting lifespan and damage propagation~\cite{harris2006, morales2019}. 
Beyond this application, virtual sensing has demonstrated its value in a broad range of fields, including intelligent buildings \cite{yoon2022virtual}, process engineering \cite{kano2013virtual, jiang2020review}, tool condition monitoring \cite{serin2020review}, and environmental monitoring~\cite{wang2022artificial}.  
Virtual sensing techniques are typically divided into \textit{model-based} or \textit{data-driven} approaches. While \textit{model-based} approaches leverage physical laws and principles to develop system models, their effectiveness is limited when complete or accurate system knowledge is  lacking~\cite{sun2021survey}.
In contrast, data-driven virtual sensing, which infers complex relationships directly from sensor data, offers a flexible alternative. It can be used to prolong the lifespan of physical sensors by providing estimates even when the original sensor fails, degrades over time, or encounters operational limitations like power depletion ~\cite{zhao2024virtual}. Additionally, data-driven models can be transferred and adapted to new systems, or even from simulations to real-world scenarios, potentially reducing the need for extensive data collection and model retraining.
Nonetheless, it faces significant challenges when applied to complex real-world scenarios posed by the heterogeneity of the sensor signals as well as the significant influence of environmental and operational conditions on the signal characteristics.

\textbf{Heterogeneous temporal dynamics.}
The increasing complexity of modern systems necessitates the deployment of multimodal sensor networks, which utilize diverse sensors to capture data across a wide range of frequencies, temporal dynamics, and spatial scales. 
For instance, in bridge health monitoring, high-frequency accelerometers are installed to capture rapid structural vibrations at high sampling rates, providing critical insights into the instantaneous changes in the bridge's structural responses. 
In contrast, low-frequency strain gauges, sampled at lower rates, are deployed to monitor gradual deformations over extended periods, revealing subtle shifts in the bridge's dynamic behavior~\cite{seo2016summary}. This highlights how different sensor modalities are strategically deployed to capture multi-scale temporal dynamics present within a single system. Furthermore, sensor modalities often exhibit varying spatial coverage, with accelerometers providing a global perspective into structural health and strain gauges offering localized measurements for locating potential damage.

\textbf{Impact of exogenous variables.}
Data-driven virtual sensing is further complicated by the influence of exogenous variables, which can be categorized as either \textit{control variables} (intentionally manipulated parameters) or \textit{environmental factors} (uncontrolled external conditions).
These variables can significantly alter a system's behavior and the temporal characteristics of sensor measurements, with modality-specific effects that can lead to shifts in signal magnitude or frequency.
For instance, in a rotating machine, increasing the rotational speed (a control variable) may lead to higher temperatures and higher frequency vibrations, as captured by thermocouples and accelerometers. 
Similarly, in bridge health monitoring, ambient temperature fluctuations (an environmental factor) can induce thermal stresses, affecting strain gauge readings and shifting the frequency spectrum of acceleration data captured by vibration sensors~\cite{borah2021effect}.

The heterogeneity in temporal dynamics, coupled with the influence of exogenous variables, poses significant challenges for traditional data-driven methods.
For instance, CNNs are well-suited for capturing local patterns but often struggle to model long-range dependencies across vastly different time scales. Conversely, RNNs are designed to capture long-range temporal dependencies but may not effectively model high-frequency signals or extract localized features. 
While previous approaches, such as the multi-scale CNN~\cite{yuan2020soft} and multiscale attention-based models~\cite{yuan2024quality}, have attempted to incorporate multi-scale analyses, they primarily focus on modeling lagged correlations across sensors and extracting features at different temporal scales. 
However, they often do not explicitly account for the diverse temporal dynamics present in the data or the influence of exogenous variables. 
This limitation becomes critical as the complexity of systems and the variety of sensors increase.

Graph Neural Networks (GNNs) offer a promising approach to tackling these challenges. 
By representing sensors as nodes and their relationships as edges, GNNs can explicitly model the spatial dependencies within the network, a key aspect often overlooked by traditional methods.
Through message-passing techniques, GNNs can iteratively exchange and aggregate information between nodes, building a global understanding of the system's state from local interactions~\cite{battaglia2018relational}. 
Their inherent ability to model spatial-temporal dependencies~\cite{jin2023survey} makes GNNs well-suited for capturing the complex interactions between sensors over time. GNNs have shown promise in virtual sensing, with recent work demonstrating their use in estimating values at unmeasured locations~\cite{de2024graph} and incorporating domain knowledge for improved accuracy~\cite{niresi2024physics}.

However, current GNN approaches often fall short in handling the heterogeneous nature of real-world sensor data.
These methods typically assume that all nodes in the graph represent sensors with similar signal characteristics, such as sampling rates and temporal dynamics. 
While some GNNs have been applied to heterogeneous sensor networks with different sensor types (e.g., temperature, humidity, pressure), these scenarios often still involve signals with relatively similar temporal characteristics~\cite{jin2023survey}.
Moreover, existing GNNs struggle to account for the diverse impact of exogenous variables on different sensor modalities. The resulting shifts in signal magnitude or frequency can lead to inaccurate predictions, particularly under varying operating conditions. This limitation becomes pronounced when dealing with sensors that have different sensitivities to environmental factors or operational changes, as the GNNs may not be able to effectively model these variations.

To overcome these limitations, we propose a novel Heterogeneous Temporal Graph Neural Network (HTGNN) framework, specifically designed for virtual sensing in complex industrial environments.
By explicitly modeling signals with distinct temporal dynamics as separate node types, and incorporating operating condition context, our HTGNN can effectively fuse information from diverse sensor modalities and account for the varying influence of exogenous variables. 
This enables more accurate prediction of essential parameters for Prognostics and Health Management (PHM), overcoming the limitations of current data-driven virtual sensing methods. To the best of our knowledge, HTGNN presents the first design of such an architecture specifically designed to analyze such diverse sensor modalities and account for exogenous variables in virtual sensing.
Specifically, this work makes the following significant contributions to data-driven virtual sensing for complex industrial systems:
\begin{itemize}[itemsep=5pt,topsep=5pt,parsep=0pt,partopsep=0pt]
\item \textbf{Heterogeneous interaction modeling}: 
HTGNN models both \textit{intra-modality} and \textit{inter-modality} interactions within heterogeneous sensor networks in distinct ways.
\item \textbf{Operating condition-aware dynamics}: 
\revtext{We introduce new general purpose modules to} capture the unique temporal dynamics of each sensor modality under varying operating conditions, accounting for the influence of exogenous variables.
\item \textbf{\revtext{First heterogeneous} datasets for load virtual sensing}: 
We \revtext{contribute} two \revtext{new}, publicly available datasets, \revtext{specifically designed to advance research in virtual sensing with multimodal sensor data and varying operating conditions, including}: a real-world \revtext{experimental} bearing dataset and a simulated train-bridge-track interaction dataset. 
\item \textbf{Comprehensive ablation study}:  We conducted a detailed ablation study to analyze the impact of each HTGNN component on overall model performance.
\end{itemize}

The remainder of this paper is organized as follows. Sec.~\ref{sec:rw} reviews the related work.
Sec.~\ref{sec:framework} formally defines the problem and presents the HTGNN framework. Sec.~\ref{sec:casestudies} describes the case studies, while Sec.~\ref{sec:exp_design} details the experimental setup and the baseline methods. Sec.~\ref{sec:result} presents and discusses the results, and Sec.~\ref{sec:conclusion} concludes the paper with a summary of contributions and future directions.

\section{Related Works}
\label{sec:rw}
In this section, we review state-of-the-art virtual sensing methods, covering both model-based and data-driven approaches, \revtext{discuss recent advances in deep time series modeling for handling multiscale dynamics and exogenous variables, and} explore advancements in graph neural networks for time series modeling and their emerging applications in virtual sensing.

\textbf{Model-based virtual sensing} approaches leverage well-established physical laws, often formulated as first-principle models, to model system behavior and estimate unmeasured quantities. 
However, due to challenges in fully capturing real-world complexities and handling noisy sensor data, these models are rarely employed in isolation \cite{jiang2020review}. Instead, they are frequently integrated with data-driven methods like Kalman filtering and Gaussian processes.
Kalman filtering, an approach that dynamically updates system states in real-time based on noisy sensor measurements, has found diverse applications.
For instance, variants of Kalman filtering have been employed for bearing load estimation \cite{kerst2019model}, actuator health monitoring \cite{schimmack2018extended}, vehicle sideslip angle estimation \cite{bertipaglia2022two} and wind turbine load prediction~\cite{branlard2024digital}. 
Additionally, Gaussian processes, a probabilistic modeling technique, have been utilized to infer unknown load dynamics from latent force models~\cite{bilbao2022virtual}. 
Despite their effectiveness, these methods rely on extensive prior knowledge of the underlying physics, which can be challenging or infeasible to obtain in many real-world scenarios~\cite{jiang2020review, sun2021survey}.

\textbf{Data-driven virtual sensing} methods learn complex relationships directly from sensor data, bypassing the need for explicit physical models \cite{sun2021survey}. 
Autoencoders (AEs), for example, have been applied due to their ability to learn latent representations of input data. Since AEs are traditionally unsupervised, they are often modified for supervised learning tasks in virtual sensing by incorporating labeled data or modifying their architecture. 
For instance, Shen \textit{et al.} \cite{shen2020nonlinear} developed a semi-supervised probabilistic latent variable regression model using stacked Variational AE (VAE). 
Notably, VAEs perform well in learning data distribution and data imputation, making them valuable tools for addressing the challenge of missing data in virtual sensing~\cite{xie2019supervised}.
Moreover, Recurrent Neural Networks (RNNs), particularly Long Short-Term Memory (LSTM) networks, have shown effectiveness in estimating variables with dynamic characteristics. Examples include wind turbine blade root bending moments \cite{dimitrov2022virtual}, hydrocracking processes~\cite{yuan2020deep}, and indoor air quality \cite{loy2020soft, ma2023multi}.
Convolutional Neural Networks (CNNs) have also been applied in virtual sensing, particularly for data with diverse temporal patterns. This architecture has been utilized in various applications, such as mapping partial vibration measurements to structural responses~\cite{sun2017data}, and chemical process quality prediction from process variables~\cite{wang2019dynamic}.

\revtext{\textbf{Multiscale modeling and exogenous variables.} 
Virtual sensing must address the challenges of heterogeneous temporal dynamics and exogenous variable influence. Deep time series models, while often not designed as regression models for virtual sensing, have explored multiscale modeling and exogenous variable integration~\cite{wang2024deep}. 
For instance, Pyraformer~\cite{liu2022pyraformer} utilizes a pyramidal attention module to capture multi-resolution features in time series for forecasting tasks. Similarly, TimeMixer~\cite{wang2023timemixer}, an MLP-based architecture, leverages disentangled multiscale series for past extraction and future prediction. However, these methods have not been applied to virtual sensing problems.
Recently, incorporating exogenous variables into time series models has gained traction.
For instance, Zhao \textit{et al.}~\cite{zhao2024dyedgegat} incorporate exogenous variables as operating condition context to improve model robustness for anomaly detection. TimeXer~\cite{wang2024timexer} enhances transformer architectures by incorporating exogenous information through embedding layers, patch-wise self-attention, and variate-wise cross-attention. Similarly, Theiler \textit{et al.}~\cite{theiler2024integrating} integrate exogenous variables from future time steps to improve the forecasting accuracy of transformer models. 
However, these approaches do not explicitly model how exogenous variables can alter the frequency or scale of endogenous variables, a crucial aspect for accurate virtual sensing.
}

\revtext{\textbf{Graph neural networks (GNNs)} were originally developed for static graph representations, with applications in fields such as computational chemistry~\cite{gilmer2017neural} and social networks~\cite{kipf2017gcn,velivckovic2017graph}. Leveraging their ability to capture complex spatial dependencies, GNNs were later adapted to time-series data, giving rise to temporal GNNs. 
These models have rapidly expanded into the Industrial Internet of Things (IIoT)~\cite{dong2023iotreview}, where effectively model spatial-temporal dependencies in sensor networks is crucial for downstream tasks such as anomaly and fault detection~\cite{jin2023survey, zhu2022encoder}.}
For example, a GNN-based model leveraging graph attention mechanisms has shown promise in identifying anomalies within multivariate time series \cite{zhao2020multivariate}. Another approach employed sensor embeddings for graph construction and prediction-based anomaly detection \cite{deng2021graph}. Recently, a GNN model incorporating dynamic edges has further improved fault detection by capturing evolving relationships between time series, while also considering operating conditions \cite{zhao2024dyedgegat}. However, a persistent challenge remains in addressing heterogeneous temporal dynamics often present in sensor data.
Beyond anomaly detection, the inherent ability of GNNS to model interactions within sensor networks has shown promise in a board range of applications. These include predicting remaining useful life of bearing~\cite{yang2022bearing}, forecasting air quality ~\cite{ouyang2021spatial}, estimating states in hydro power plants~\cite{theiler2024graph}, and calibrating ozone sensors using low-cost sensor networks.

\textbf{GNN-based virtual sensing.} 
Despite their demonstrated potential in modeling complex systems for time series analysis, their application in virtual sensing, particularly for predicting unmeasured quantities from available sensor data, is still in its early stages.
Recent studies have leveraged GNNs for tasks like estimating values at unmeasured locations. For example, Niresi~\textit{et al.}~\cite{niresi2024physics} utilized physics-enhanced virtual nodes in GNNs for virtual sensing in district heating networks. Moreover, Felice \textit{et al.}~\cite{de2024graph} introduced graph structures to reconstruct missing sensor measurements in both weather and photovoltaic systems. 
However, these approaches often rely on simplifying assumptions that limit their applicability to complex, real-world systems. For example, they may assume homogeneous signal attributes (e.g., sampling rates and temporal dynamics) across all sensors or consider only systems with homogeneous sensor setups. Additionally, current state-of-the-art approaches often struggle to account for the diverse influence of exogenous variables on different sensor modalities. 
These limitations hinder the ability of existing GNN-based methods to accurately capture heterogeneous temporal dynamics and predict system behavior under varying operating conditions, especially in complex systems where direct measurement of certain quantities is infeasible or impractical.

\section{Proposed Framework}
\label{sec:framework}
In this paper, we establish the notation where bold uppercase letters (e.g., $\mathbf{X}$), bold lowercase letters (e.g., $\mathbf{x}$), and calligraphic letters (e.g., $\mathcal{V}$) denote matrices, vectors, and sets, respectively. Time steps are indicated by superscripts (e.g., $\mathbf{X}^t$ is the matrix $\mathbf{X}$ at time $t$), while subscripts identify specific nodes (e.g., $\mathbf{x}_i$ is the vector for node $i$).
\subsection{Problem Statement}
\label{sec:problem_statement}

We address the challenge of virtual sensing in complex systems equipped with heterogeneous sensor networks. Heterogeneity arises from the diverse nature of sensor modalities. 
This diversity is characterized by differences in the underlying physical phenomena they capture, resulting in variations in their operated frequencies (sampling rates), temporal dynamics (gradual vs. abrupt changes), and the scope of information they provide about the system (local vs. global aspects).
Specifically, our dataset consists of N sensor signals sampled at discrete time intervals. Based on their dominant frequency characteristics, these signals naturally fall into two primary types: low-frequency signals $\mathbf{x}_{L}$ and high-frequency signals $\mathbf{x}_{H}$.
At a specific time instance $t$, the low-frequency and high-frequency data are represented as:
\begin{align}
    \mathbf{x_L}^{t} &= [x_{L_1}^{t}, x_{L_2}^{t}, ..., x_{L_{N_L}}^{t}]^T \in \R^{N_L}, \\
    \mathbf{x_H}^{t} &= [x_{H_1}^{t}, x_{H_2}^{t}, ..., x_{H_{N_H}}^{t}]^T \in \R^{N_H},
\end{align}
where $N_L$ and $N_H$ are the number of each sample of data type, and $x_{L_i}^{t}$ and $x_{H_i}^{t}$ denote the measurements at time $t$ from the $i^{th}$ sensor for low-frequency and high-frequency signals, respectively.
To align the data for analysis despite varying sampling frequencies, common strategies include upsampling the low-frequency using interpolation techniques or downsampling the high-frequency data using aggregation techniques. This allows for a unified temporal representation of the heterogeneous sensor signals.  
Although this paper presents our methodology based on this upsampling/downsampling approach, it is worth noting that the framework could also be readily extended to incorporate the alternative strategy of employing a longer observation window for the low-frequency data to capture its slower dynamics while maintaining its original sampling rate.

Additionally, we incorporate a set of $N_w$ exogenous variables $\mathbf{w}^{t} \in \mathbb{R}^{N_w}$, which can represent control inputs or external factors influencing the system's behavior and its internal states. However, the exogenous variables are not affected by the system's internal states.
Assuming the low-frequency data has been upsampled to match the higher sampling rate of the high-frequency data, we construct time-series samples for each sensor modality using a sliding window of length $L$, resulting in the following representations:
\begin{align}
\mathbf{X_L}^{t_l:t} &= [\mathbf{x_L}^{t_l}, \cdots, \mathbf{x_L}^{t-1}, \mathbf{x_L}^{t}] \in \mathbb{R}^{N_L \times L},\\
\mathbf{X_H}^{t_l:t} &= [\mathbf{x_H}^{t_l}, \cdots, \mathbf{x_H}^{t-1}, \mathbf{x_H}^{t}] \in \mathbb{R}^{N_H \times L},\\
\mathbf{W}^{t_l:t} &= [\mathbf{w}^{t_l}, \cdots, \mathbf{w}^{t-1}, \mathbf{w}^{t}] \in \mathbb{R}^{L},
\end{align}
where $t_l = t - L + 1 > 0$ is the the starting point of the observation window.

Our objective is to develop a virtual sensor, represented by a function 
\begin{equation}
    f(\mathbf{X_L}^{t_l:t}, \\\mathbf{X_H}^{t_l:t}, \mathbf{W}^{t_l:t})= \mathbf{y}^{t},
\end{equation}
that accurately estimates the target variable $\mathbf{y}^t \in \R^d$ at time $t$, given the heterogeneous sensor data $\mathbf{X_L}^{t_l:t}$, $\mathbf{X_H}^{t_l:t}$, and $\mathbf{W}^{t_l:t}$.
This task is challenging due to the inherent differences between low-frequency and high-frequency signals, which capture distinct aspects of the system's state.
Low-frequency signals (e.g., temperature, deformation) typically reflect gradual changes, while high-frequency signals (e.g., acoustic signals, acceleration) reveal immediate mechanical interactions and anomalies. 
These differences in frequency not only affect the data processing strategy but also the information fusion of these signals in real-time condition monitoring. 
Moreover, the temporal dynamics of these signals can be significantly affected by varying operating conditions (e.g., changing load, speed, or environmental factors), making accurate virtual sensing a challenging problem. 
In the following sections, we will detail our approach to address these challenges and develop a robust virtual sensor for reliable virtual sensing under varying operation conditions.

\subsection{Framework Overview}
\begin{figure*}[tbhp]
  \centering
    \includegraphics[width=1.\linewidth]{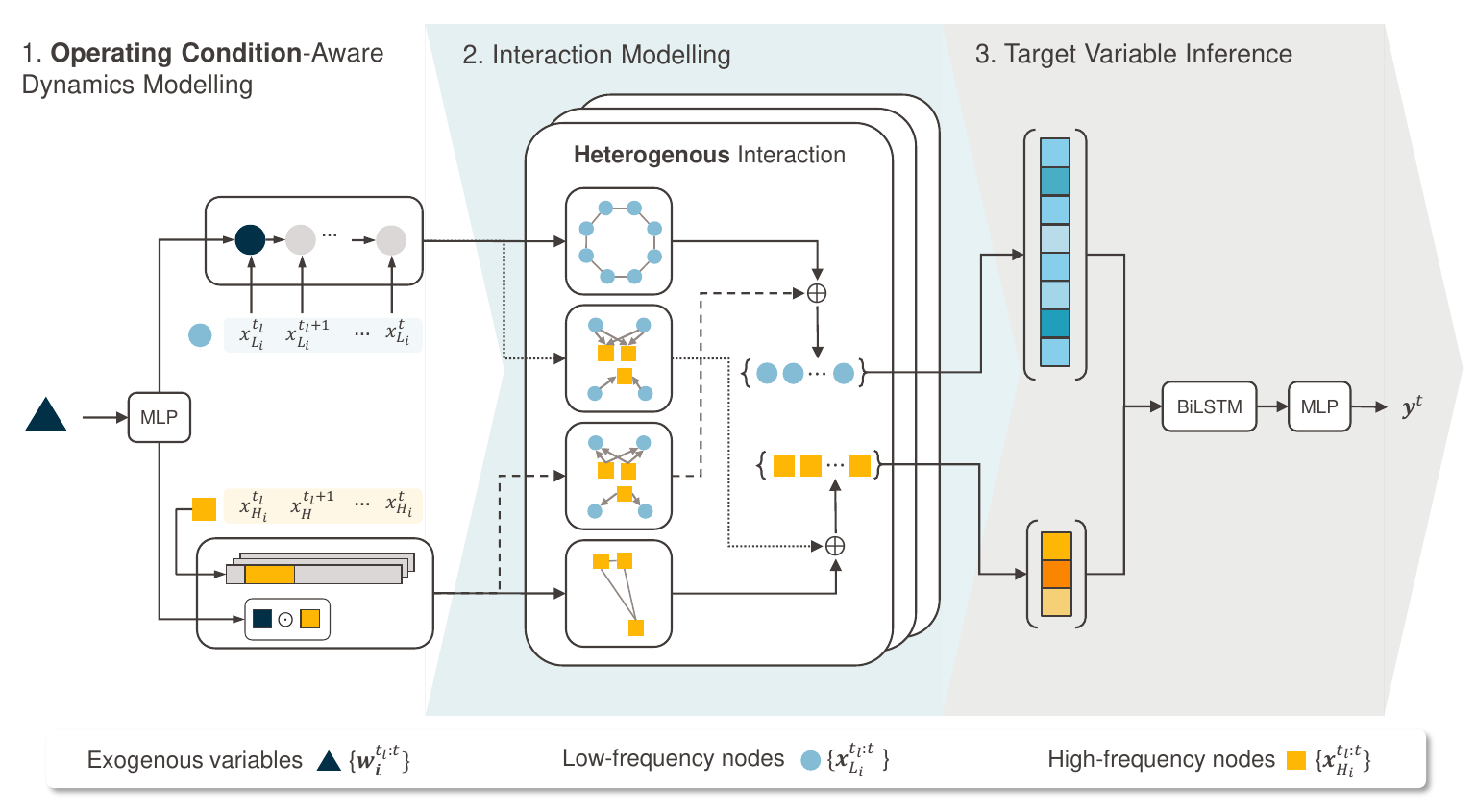}
  \caption{Architecture of the proposed Heterogeneous Temporal Graph Neural Network (HTGNN) for 
  \revtext{virtual sensing. HTGNN effectively infers target variable $\mathbf{y}^{t}$ by capturing complex relationships between heterogeneous sensor nodes with diverse modalities, including low-frequency signals ($\mathbf{X_L}^{t_l:t}$, blue circles), high-frequency signals ($\mathbf{X_H}^{t_l:t}$, orange squares) and exogenous variables ($\mathbf{W}^{t_l:t}$, dark blue triangle).
  The model incorporates three key stages: \textbf{(1) Operating condition aware dynamics modeling}: A Multilayer Perceptron (MLP) extracts operating conditions, informing both a Gated Recurrent Unit (GRU) for low-frequency dynamics and a gated Convolutional Neural Network (CNN) for high-frequency dynamics extraction, capturing operating condition-aware temporal patterns within each signal type. \textbf{(2) Heterogeneous interaction learning}: Specialized Graph Neural Networks (GNNs) learn complex inter- and intra-modality interactions between heterogeneous sensor nodes. \textbf{(3) Target variable inference}: A Bidirectional Long Short-Term Memory (BiLSTM) network followed by an MLP analyzes the combined information from the graph embedding to predict the target variable.
  }}
  \label{fig:htgnn}
\end{figure*}
We propose a novel Heterogeneous Temporal Graph Neural Network (HTGNN) for real-time virtual sensing of load in complex systems. 
Our proposed algorithm learns a virtual sensing function, $f(\mathbf{X_L}^{t_l:t}, \mathbf{X_H}^{t_l:t}, \mathbf{W}^{t_l:t})= \mathbf{y}^{t}$, which maps low-frequency signals $\mathbf{X_H}^{t_l:t}$,  high-frequency signals $\mathbf{X_H}^{t_l:t}$,  and exogenous variables $\mathbf{W}^{t_l:t}$ within a time window $[t_l,t]$ to the corresponding load $\mathbf{y}^t$ at time $t$. 
Although we primarily distinguish between low- and high-frequency signals in this function, it is important to note that the heterogeneity extends beyond mere frequency differences, encompassing variations in temporal dynamics and spatial scales as well. High-frequency signals often capture instantaneous changes and global information about the system, while low-frequency signals typically reflect gradual changes and more localized information.

Unlike conventional approaches that treat all sensors identically, HTGNN explicitly models the distinct temporal dynamics of different sensor modalities as well as intra- and inter-modality interactions.
By representing different sensor modalities as distinct node types within an aggregated temporal graph, the HTGNN  allows for the extraction of unique temporal dynamics specific to each sensor modality.  Specialized GNNs then model the interactions between these modalities, enhancing the accuracy of the target variable inference. 
This approach provides a substantial improvement over conventional homogeneous temporal GNN methods, which typically handle only a single type of sensor relation.
Fig.~\ref{fig:htgnn} illustrates the HTGNN's architecture, which consists of the following key components:

\begin{itemize}[itemsep=5pt,topsep=5pt,parsep=0pt,partopsep=0pt]
\item \textbf{Heterogeneous Temporal Graph Construction (Sec.~\ref{sec:heteo_graph}):} A graph representing the connectivity of the sensor network, incorporating diverse sensor modalities (i.e., high-frequency and low-frequency signals), is constructed.

\item \textbf{Context-Aware Heterogeneous Dynamics Extraction (Sec.~\ref{sec:htgnn_node_dynamcics}):} Unique temporal patterns inherent to each sensor modality are extracted using dedicated encoders, explicitly accounting for specific operating conditions and individual signal characteristics.

\item \textbf{Heterogeneous Interaction Modeling (Sec.~\ref{sec:htgnn_interaction}):} Tailored GNN architectures are employed to effectively model the diverse interaction types present within the complex sensor networks.

\item \textbf{Target Variable Inference (Sec.~\ref{sec:htgnn_load_prediction}):} A BiLSTM layer is employed to infer the target variable, selectively attending to and weighing information from various nodes and features within the learned graph.
\end{itemize}

In the subsequent sections, we provide a detailed explanation of each component of the HTGNN framework. We will illustrate how HTGNN effectively addresses the challenges of virtual sensing in complex environments characterized by heterogeneous temporal dynamics.

\subsection{Heterogeneous Temporal Graph Construction}
\label{sec:heteo_graph}
\textbf{Heterogeneous Static Graph.} As outlined by \cite{shi2022heterogeneous}, a Heterogeneous Static Graph (HSG) is defined as $\mathcal{G}=(\mathcal{V},\mathcal{E})$, comprising a set of nodes $\mathcal{V}$ and a set of edges $\mathcal{E}$. Both nodes and edges in this graph can belong to various types. The graph features a node-type mapping function $\phi: \mathcal{V} \rightarrow \mathcal{A}$ and an edge-type mapping function $\psi: \mathcal{E} \rightarrow \mathcal{R}$, where $\mathcal{A}$ and $\mathcal{R}$ represent the respective sets of node and edge types. This setup ensures a diverse graph structure, with the combined variety of node and edge types exceeding two, i.e., $|\mathcal{A}| + |\mathcal{R}| > 2$.

\textbf{Heterogeneous Temporal Graph.} Building upon the Heterogeneous Static Graph (HSG), a Heterogeneous Temporal Graph (HTG) is defined as a sequence of HSGs across  $T$ time steps,  represented as $\mathcal{G}^{T} = \{\mathcal{G}^{t_1}, \ldots, \mathcal{G}^{t_T}\}$. Each graph within this series, $\mathcal{G}^t = (\mathcal{V}^t, \mathcal{E}^t)$, captures  the state of the graph at a specific time $t$. Both the node and edge type mapping functions, $\phi$ and $\psi$, are maintained consistently across these time steps. The HTG is aggregated into a unified form as follows:
\begin{equation}
\mathcal{G}^{T}= \left(\bigcup_{t=t_1}^{t_T} \mathcal{V}^t, \bigcup_{t=t_1}^{t_T} \mathcal{E}^t\right),
\end{equation}
which integrates nodes $\mathcal{V}^t$ and edges $\mathcal{E}^t$ from all time steps, thereby preserving the heterogeneity defined by $\phi$ and $\psi$. This structure allows for the analysis of dynamic interactions and changes over time while maintaining the diverse characteristics of the nodes and edges.

\textbf{Heterogeneous sensor graph construction.} To effectively model the heterogeneous sensor signals from a heterogeneous sensor network, we construct an HTG. This graph consists of two types of nodes: low-frequency (L) nodes characterized by attributes $\mathbf{X_L}^{t_l:t}$, and high-frequency (H) nodes characterized by attributes $\mathbf{X_H}^{t_l:t}$. The edges in the graph are defined by the relationships between these node types, specifically  L-L, H-H, L-H, and H-L, and are assumed to be consistent over time. This framework enables the HTG to capture and analyze the interactions and temporal evolution between low-frequency and high-frequency signals effectively. A detailed visualization of the HTG is provided in Fig.~\ref{fig:htgnn}, illustrating the dynamic interplay of sensor types within the network.

\subsection{Context-aware Node Dynamics Extraction}
\label{sec:htgnn_node_dynamcics}
In complex systems, the behavior of individual sensor nodes is significantly influenced by the global operating context, which includes various exogenous variables.  Changes in these variables, such as control inputs and environmental conditions, can significantly impact both the magnitude and the frequency characteristics of the sensor signals.  For instance, for low-frequency signals, shifts in exogenous variables may alter signal magnitude, such as increased deformation due to higher temperatures. Conversely, high-frequency signals may experience changes in both magnitude and frequency content, such as increased vibration amplitude and frequency associated with higher rotational speed in bearings.

To account for these important influences, our HTGNN model leverages context-aware dynamics extraction for each node. This approach extends the strategy proposed in~\cite{zhao2024dyedgegat} to explicitly model shifts in both magnitude and frequency.
We extract contextual information from exogenous variables and integrate it into the dynamics modeling of both low-frequency and high-frequency sensor modalities using specialized techniques. This enables the HTGNN to effectively capture how the operating context influences the temporal behavior of each sensor, ultimately leading to more accurate and reliable virtual sensing.

\subsubsection{Encoding Exogenous Variable.}
To extract a meaningful representation of the operating context from the exogenous variables (e.g., rotational speed, ambient temperature), we employ a Multi-Layer Perceptron (MLP). Assuming that these variables exhibit relatively minor fluctuations within a specific observation window, we first calculate the average value $\bar{w}^{t_l:t} = \frac{1}{L} \sum_{i=t_l}^{t} w^i \in \R$  over the selected time window $[t_l: t]$. The MLP then processes this average value to generate a hidden representation $\mathbf{h}_w \in \R^{d_w}$ of the operating context. This representation is used to augment the dynamics extraction from other sensor modalities. Formally, the process can be expressed as:
\begin{equation}
\mathbf{h}_w = \text{MLP}(\bar{w}^{t_l:t}).
\label{eq:exo_mlp}
\end{equation}
This approach ensures that the model captures the essential contextual information, enhancing the accuracy and reliability of the virtual sensing system.

\subsubsection{Encoding Low-Frequency Signals.} 
\label{sec:htgnn_rnn_encoder}
\begin{figure}[tbhp]
  \centering
    \includegraphics[width=.5\linewidth]{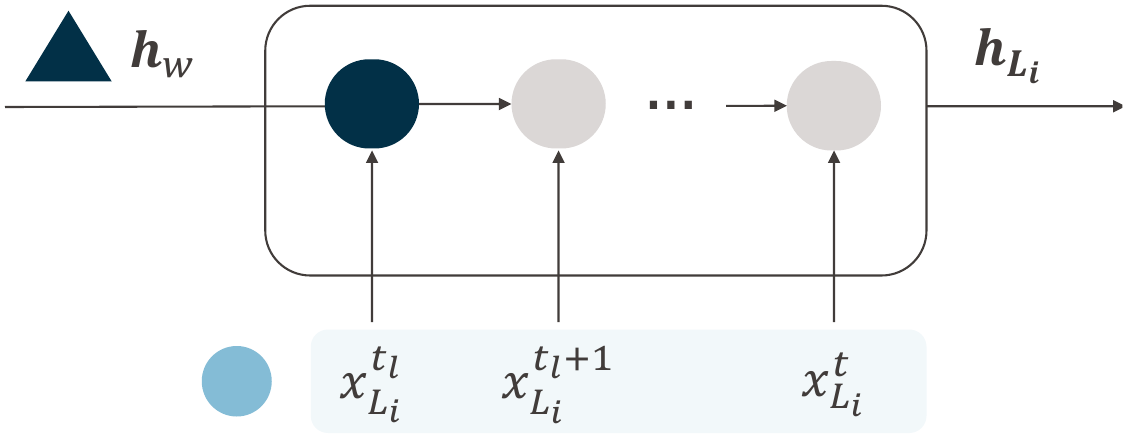}
  \caption{Architecture of a Gated Recurrent Unit (GRU)-based low-frequency signal encoder with exogenous variable encoding as the initial state.}
  \label{fig:htgnn_encoder_gru}
\end{figure}
In complex systems, the magnitude of low-frequency signals is often directly influenced by operating conditions. For instance, higher temperatures may induce increased thermal expansion, resulting in more significant deformation of structures such as bridges. This deformation is then manifested as higher values in measurements on strain gauges. Such inherent correlations between low-frequency signals and operating conditions underscore the critical need to incorporate contextual information from the operating and environmental conditions into the dynamics modeling process.

To effectively model the relationship between low-frequency signals and operating conditions, we utilize a Gated Recurrent Unit (GRU) network. 
Building on our previous work~\cite{zhao2024dyedgegat}, we initialize the GRU's hidden state with the exogenous variable encoding $\mathbf{h}_w$ from Eq.~\ref{eq:exo_mlp}. This initialization allows the dynamics encoder to immediately integrate this contextual information when processing the low-frequency signal sequence $\mathbf{x_L}_j^{t_l:t}$. This approach is particularly effective because the context of operating conditions often imposes a bias on the low-frequency signals, affecting their magnitude and overall trend. By embedding this contextual bias as the initial state, the GRU is better equipped to accurately model the temporal evolution of the signal.
For each low-frequency sensor node $i$, the GRU updates its cell state at each time step $\tau$ as follows: 
\begin{equation}
    {\mathbf{h_{L}}}_i^{\tau} = \text{SiLU}\left(\text{GRU-Cell}(\mathbf{x_L}_i^{\tau}, \mathbf{h_L}_i^{\tau-1})\right), \forall \tau \in [t_l, t].
    \label{eq:low_freq_node_gru}
\end{equation}
We use the final state $\mathbf{h_L}_i^t \in \mathbb{R}^{d_T}$, which encapsulates the encoded dynamics of node $i$ up to time $t$ along with the operational state context, as the low-frequency node representation $\mathbf{h_L}_i \in \mathbb{R}^{d_L}$.

\subsubsection{Encoding High-Frequency Signals.}  
\begin{figure*}[tbhp]
  \centering
    \includegraphics[width=.85\linewidth]{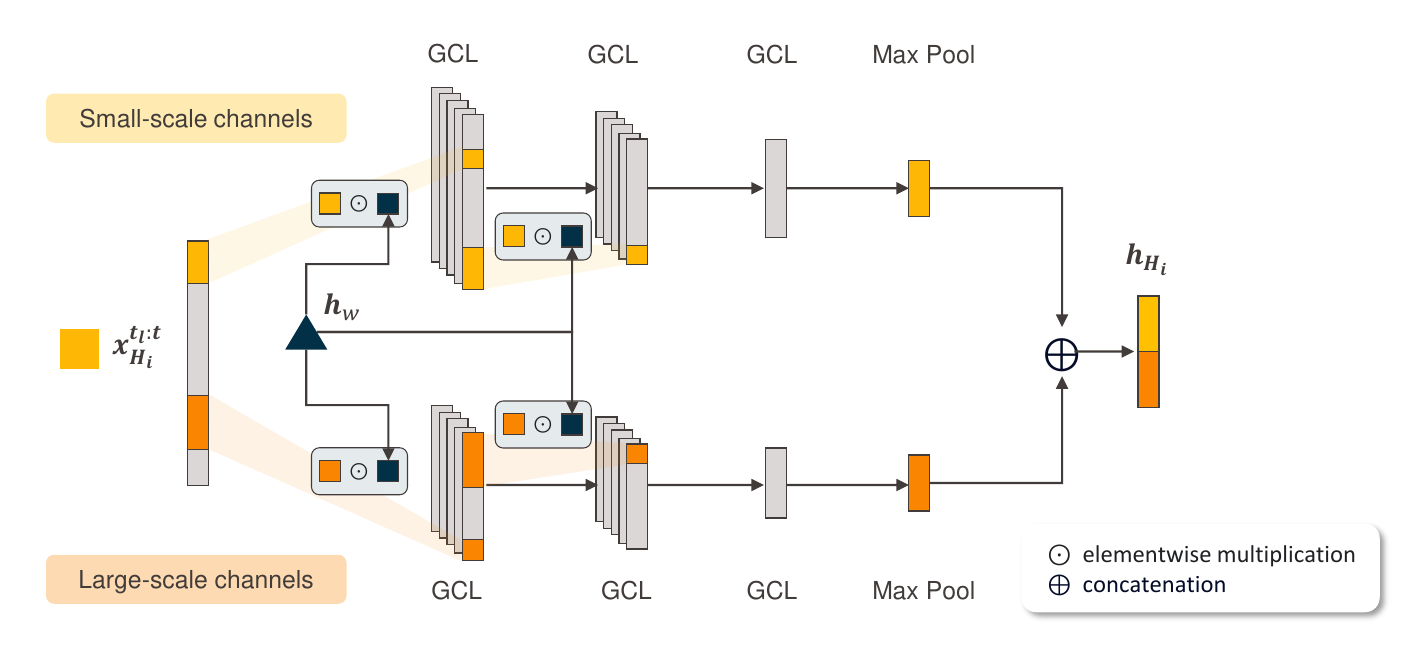}
  \caption{Architecture of a multi-scale Gated Convolutional Layers (GCLs)-based encoder for high-frequency signals, considering exogenous variable encoding as the gating signal.}
  \label{fig:htgnn_encoder_cnn}
\end{figure*}
\label{sec:htgnn_gcnn_encoder}
Changes in operating conditions can significantly affect both the magnitude and frequency characteristics of high-frequency signals in complex systems. For instance, an increase in rotational speed within a bearing can cause the dominant frequencies in the vibration signal to shift towards higher magnitudes and frequencies. 
To address these shifts in high-frequency signals, we use one-dimensional Convolutional Neural Networks (1D-CNNs), which are effective in capturing patterns within time-series data~\cite{kiranyaz20211d}. However, standard 1D-CNNs may struggle to adapt to changes in signal frequency caused by varying operating conditions.
To overcome this limitation, we propose a 1D Gated Convolutional Neural Network (1D-GCNN). This model integrates a gating mechanism into the traditional 1D-CNN architecture, enabling dynamic adjustment of its frequency focus based on the operating context. This gating mechanism allows the model to identify and prioritize the frequency components most relevant under varying conditions, resulting in more robust feature extraction.

\textbf{1D Gated Convolutional Layer (GCL)}.
The core component of our 1D-GCNN is the Gated Convolutional Layer (GCL). It operates on an input sequence $\mathbf{x}_i\in \R^{d_x}$ and the exogenous variable encoding $\mathbf{h_w}$, functioning as follows:
\begin{align}
\mathbf{z}^t &= \text{Conv1D}(\mathbf{x}_i, \mathbf{W_z}) \\
\mathbf{g}^t &= \sigma(\mathbf{W_g} \mathbf{h_w} + \mathbf{b_g}) \\
\mathbf{o}^t &= \mathbf{z}^t \odot \mathbf{g}^t
\label{eq:gated_conv_layer}
\end{align}
In this framework, the gating signal $\mathbf{g}^t\in \R^{d_o}$ functions as a dynamic filter, modulating the importance of different frequency components in the convolution output $\mathbf{z}^t\in \R^{d_z}$. This modulation is achieved through element-wise multiplication, which applies the gating effect independently to each feature, allowing for tailored frequency responses based on the current operating context. The learnable parameters in this model include the weight matrices $\mathbf{W_z} \in \R^{d_x\times d_z}$, $\mathbf{W_g}\in \R^{d_z\times d_o}$, and the bias $\mathbf{b_g}\in \R^{d_o}$, enabling adaptability and fine-tuning of the gating mechanism.

\textbf{Multi-scale frequency encoding}. 
High-frequency signals often exhibit complex multi-scale characteristics, including rapid fluctuations and slower modulations that may vary according to the operating context. To accurately capture this multi-scale nature, we employ two parallel stacks of GCLs, each focusing on different temporal scales, each focusing on different temporal scales.
The first stack, 1D-GCNN\textsubscript{small}, prioritizes the extraction of small-scale, high-frequency features using smaller kernels and dilations. The second stack, 1D-GCNN\textsubscript{large}, targets larger-scale lower-frequency components with larger kernels and dilations. Both stacks are conditioned on the exogenous variable encoding $\mathbf{h_w}$ (Eq.~\ref{eq:exo_mlp}), facilitating context-aware feature extraction:
\begin{align}
{\mathbf{h_{H}}_{,\text{small}, i}}^t &= \text{SiLU} \left( \text{1D-GCNN}_{\text{small}}(\mathbf{x_H}_i^{t_l:t}, \mathbf{h}_w) \right), \\
{\mathbf{h_{H}}_{,\text{large}, i}}^t &= \text{SiLU} \left(\text{1D-GCNN}_{\text{large}}(\mathbf{x_H}_i^{t_l:t}, \mathbf{h}_w) \right).
\label{eq:high_freq_node_cnn}
\end{align}
These multi-scale representations are subsequently concatenated to form the complete node representation: $\mathbf{h_{H}}_i = \left[{\mathbf{h_{H}}_{,\text{small}, i}}^t \parallel {\mathbf{h_{H}}_{,\text{large}, i}}^t\right] \in \R^{d_H}$. This integrated representation effectively encapsulates both the intrinsic dynamics of the signal and the influence of external operating conditions.

\subsection{Heterogeneous Interaction Modelling}
\label{sec:htgnn_interaction}
To effectively capture the complex relationships between sensor nodes and explicitly account for the influence of operating conditions on these interactions, our HTGNN model strategically models heterogeneous interactions within the temporal graph.
We address two distinct types of interactions: \revtext{intra-modality} interactions among nodes with similar signal characteristics  (e.g., low-frequency to low-frequency, high-frequency to high-frequency) and \revtext{inter-modality} across nodes with different signal characteristics (e.g., low-frequency to high-frequency). This approach to interaction modeling leverages the node dynamics previously extracted in Sec.~\ref{sec:htgnn_node_dynamcics} (low-frequency nodes from Eq.~\ref{eq:low_freq_node_gru}, high-frequency nodes from Eq.~\ref{eq:high_freq_node_cnn}). This method ensures that our model comprehensively integrates both intra- and inter-modality dynamics, enhancing the accuracy and relevance of the sensor data analysis while respecting the signal characteristics.

\textbf{Intra-modality interactions.} 
To effectively capture the interdependencies among sensors with similar frequency characteristics, we employ Graph Convolutional Networks (GCNs)~\cite{kipf2017gcn}. This methodology enhances node representations by aggregating information from neighboring nodes that display correlated behaviors.  
For instance, temperature sensors in close proximity often exhibit synchronized patterns, and using a GCN allows us to exploit these correlations to refine each sensor's node representation.
The interaction between nodes is quantified through messages passed from node $j$ to node $i$, specifically within each relationship type  $r_s$ from the set  $\in \mathcal{R}_{\text{s}}$ that connects nodes of the same type. The message passing formula is given by:
\begin{equation}
m_{j \rightarrow i}^{(l, r_s)} = \frac{1}{\sqrt{\hat{d}_i} \sqrt{\hat{d}_j}} \mathbf{W}_{\phi(j),r_s}^{(l)} \mathbf{h}_j^{(l)}, \forall r_s \in \mathcal{R}_{\text{s}}, \phi(j) = \phi(i),
\end{equation}
where $\hat{d}_i$ and  $\hat{d}_j$ represent normalized node degrees, ensuring that the message weighting considers the local topology of each node, thereby stabilizing the learning process across different node densities. $\mathcal{R}_{\text{s}}$ is the set of edge types connecting nodes of the same type.

\textbf{Inter-modality interactions.} 
To capture the influence of one signal type on another within the sensor network (e.g., the impact of low-frequency signals on high-frequency signals), we employ Graph Attention Networks v2 (GATv2) \cite{brody2022gatv2}. 
This approach enables dynamic computation of attention-weighted messages, allowing the model to dynamically assess the relevance of neighboring nodes based on their interactions. The attention coefficients $\alpha_{ji}^{(l,r_d)}$ for a target node $i$ receiving a message from node $j$ under  relation $r_d \in\mathcal{R}_d$ are calculated  as follows:
\begin{equation}
\alpha_{ji}^{(l,r_d)} =\\ \text{softmax}_j \left(\mathbf{a}_{r_d}^{(l)T} \text{LeakyReLU}( \mathbf{W}_{r_d}^{(l)} \cdot [ \mathbf{h}_i^{(l)} \parallel \mathbf{h}_j^{(l)} ]) \right),
\end{equation}
where $r_d \in\mathcal{R}_d$ represents the set of edge types that connect nodes of different types. The messages are then computed as:
\begin{equation}
m_{j \rightarrow i}^{(l, r_d)} = \alpha_{ji}^{(l,r_d)} \mathbf{W}_{\phi(j),r_d}^{(l)} \mathbf{h}_j^{(l)}, \forall r_d \in \mathcal{R}_{\text{d}}, \phi(j) \neq \phi(i),
\end{equation}
\textbf{Aggregation and update:} After  aggregating messages from both same-type and different-type connections, the node representations are updated in the following manner:
\begin{equation}
\mathbf{h}_{\phi(i)}^{(l+1)} = \text{SiLU} \left( \sum_{r \in \mathcal{R}_s \cup \mathcal{R}_d} \frac{1}{|\mathcal{N}_{r}(i)|} \sum_{j \in \mathcal{N}_{r}(i)} m_{j \rightarrow i}^{(l, r)} \right).
\end{equation}
The updated node state contains incoming messages across all relationships, effectively updating node representations to reflect both homogeneous and heterogeneous influences. This enhances the model's ability to interpret complex sensor network dynamics, providing a comprehensive understanding of the system's behavior under varying operating conditions.

\subsection{Target Variable Inference}
\label{sec:htgnn_load_prediction}
After extracting the context-aware dynamics from each node, we proceed to integrate these heterogeneous node representations to infer the target variable. 
First, we flatten the final node representations of both low-frequency and high-frequency nodes into a single input vector. This vector is subsequently processed by a Bidirectional Long Short-Term Memory (BiLSTM) network. 
This novel application of a BiLSTM to fuse diverse node representations into a sequential format constitutes a unique approach for integrating heterogeneous sensor information in virtual sensing.
The BiLSTM leverages flattened node embeddings to capture dependencies and relationships between nodes and their attributes in both forward and reverse temporal directions. This bidirectional processing enables the model to gather comprehensive information across the entire graph. Unlike simpler aggregation methods such as mean or max pooling, which only capture basic global statistics, the BiLSTM is capable of selectively attending to and weighing information from various nodes and features. This selective attention significantly enhances the model's ability to learn complex relationships between sensor signals and the target variable. The final output of the BiLSTM is then passed through an MLP to generate the final prediction for the target variable $\hat{\textbf{y}}^t$.

\section{Case Studies}
\label{sec:casestudies}
To demonstrate the effectiveness of our HTGNN model in addressing the challenges posed by heterogeneous temporal dynamics and varying operating conditions, we conducted two case studies focusing on real-world systems: bearing load prediction and bridge live load prediction. 
In both cases, the systems are monitored by heterogeneous sensors, but the nature of the challenges they present differs significantly. The bearing case study highlights the impact of varying \textit{control inputs}, specifically rotational speed, on the temporal dynamics of temperature and vibration sensor signals. In contrast, the bridge case study focuses on the influence of an \textit{environmental factor}, temperature, on displacement and acceleration sensor readings. By evaluating these distinct scenarios, we aim to demonstrate the versatility of the HTGNN framework in handling diverse sensor modalities, adapting to dynamic environments, and accurately estimating critical parameters under both controlled and uncontrolled conditions.  These case studies also underscore the generalizability of our approach across different industrial and infrastructure systems, showcasing its potential for broad applicability in real-world virtual sensing applications. 

\subsection{Case Study 1: Bearing}
\label{sec:case_study_bearing}
In our initial case study, we assess the applicability of the proposed HTGNN model for estimating bearing loads under various operating conditions, using data from temperature and vibration sensors.  These sensor modalities provide complementary information: vibration is indicative of the magnitude of the load, while temperature reveals the spatial distribution of the load within the bearing. Accurate load estimation is valuable for implementing predictive maintenance strategies in industrial applications, as 
it enables operation optimization, misalignment detections~\cite{widner1976bearing}, faults diagnosis, as well as lifespan and damage propagation prediction~\cite{harris2006, morales2019}. 
Directly measuring bearing loads during operation can be intrusive, expensive, or impractical due to battery limitations~\cite{baggerohr2023}. Therefore, our approach focuses on inferring load values from readily available sensor signals, making the process more feasible and cost-effective.

\begin{figure}[tbhp]
    \centering
    \subfloat[Test-rig setup.]{
            \includegraphics[width=0.3\linewidth]{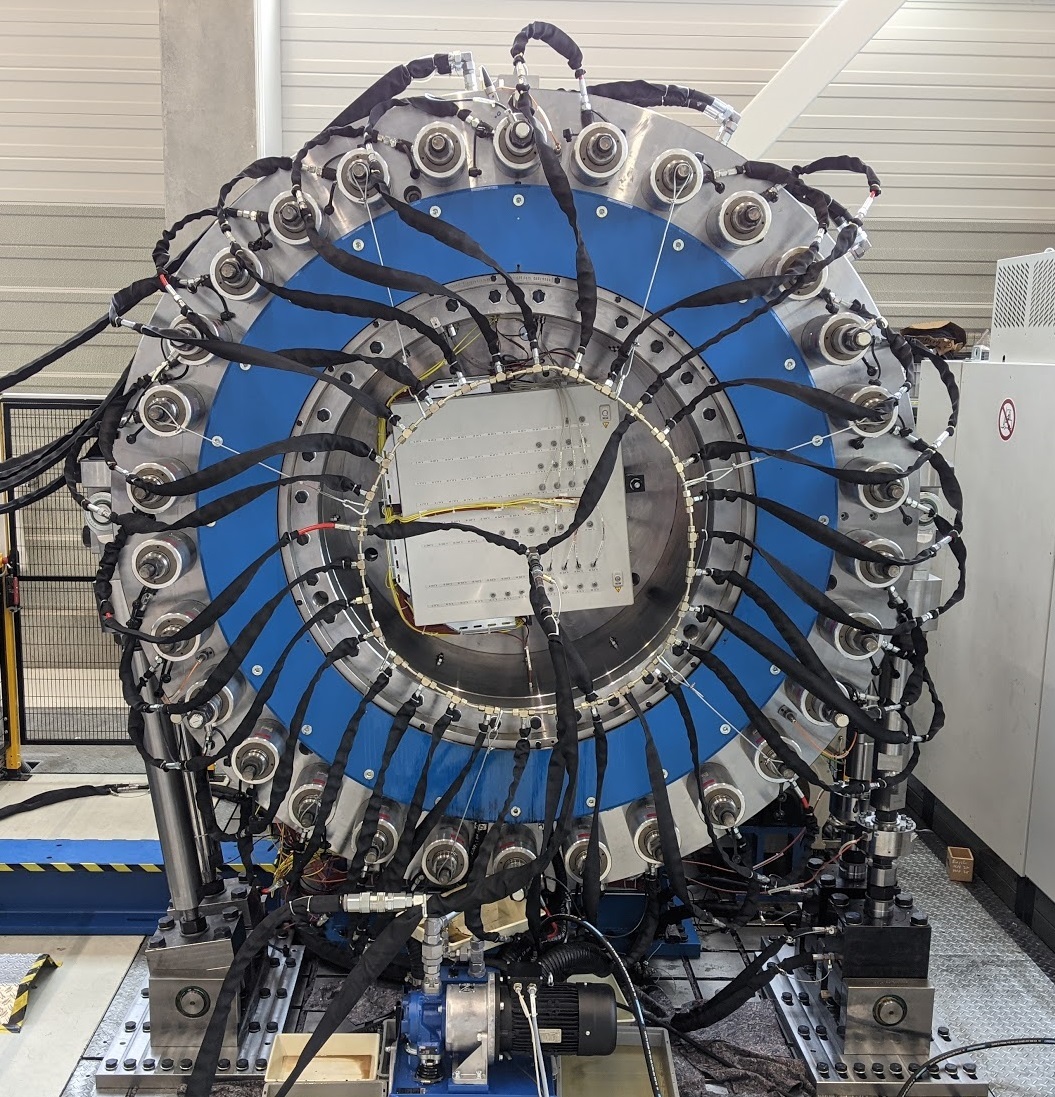}}
    \subfloat[Sensor installation locations on the test-rig.]{
        \includegraphics[width=.6\linewidth]{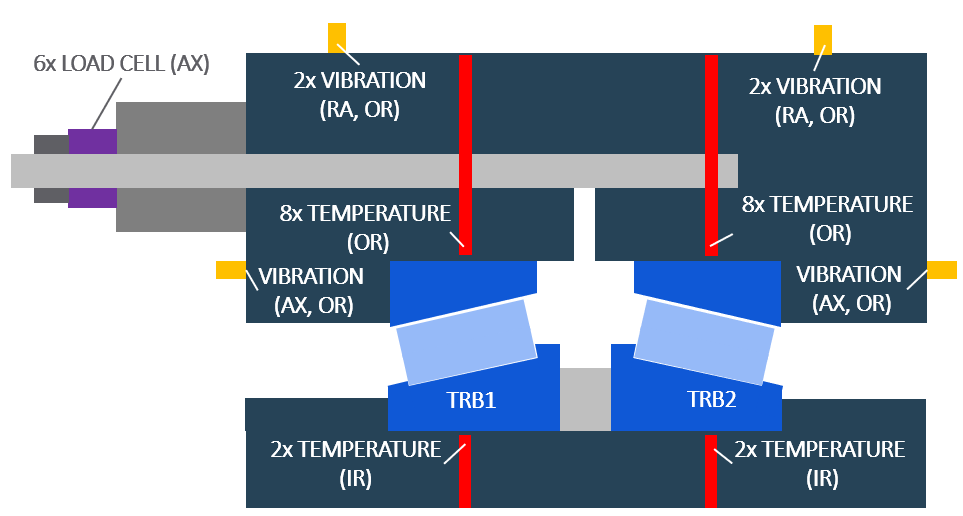}}    
    \caption{The SKF Sven Wingquist Test Centre (SWTC) TRB bearing test-rig (a) with sensor installation locations (b) for vibration, temperature, and load measurements.
}
    \label{fig:testrig_setup}
\end{figure}

\subsubsection{Bearing Test Rig Data Acquisition}
The data used in this study was collected at the SKF Sven Wingquist Test Centre (SWTC) using a face-to-face test rig with two identical single-row tapered roller bearings (TRBs). The TRBs feature a rotating inner ring, an outer diameter of 2,000 mm, an inner diameter of 1,500 mm, and a width of 220 mm, each incorporating 50 rollers. 
This setup aims to assess load conditions under various operational scenarios. 
Fig.~\ref{fig:testrig_setup} illustrates the sensor positioning on both identical TRBs. Ten temperature sensors are positioned on each bearing, with eight uniformly distributed on the outer ring (OR), and two on the inner ring (IR). Additionally, six vibration sensors on the outer ring measure both axial (AX) and radial (RA) vibrations, with sensors placed at the top and bottom of the bearing housing for the radial direction. 

Temperature is recorded at a 1 Hz sampling rate with a precision of 0.05°C. Vibration data is aggregated to 1 Hz using  Root Mean Square (RMS) aggregations. Axial and radial forces are measured and controlled by several load cells, with aggregated load values in both directions used as ground truth for this study (note that the radial load cells are not shown in the figure).


\subsubsection{Data Preprocessing}
To prepare the data for virtual sensor emulation, we first filtered the raw sensor data, retaining only cases with at least 20 minutes and up to 1 hour of continuous, stationary operation. This resulted in a final dataset comprising 164,948 samples.
To reduce noise and transient fluctuations in the temperature data, we apply a moving average filter with a 1-minute window. We focus on the rate of temperature change because the bearing temperature responds gradually to changes in load and speed. We calculated this rate over 5-minute periods to align with typical operational changes. This approach allows our model to identify the immediate impact of load changes on temperature, rather than the cumulative effects of historical variations.
After preprocessing, we split both temperature and vibration signals using a sliding window, with a length of 30 and a stride of 1.

To emulate the deployment and validation of a virtual sensor, the dataset was split by time while maintaining a representation of all 55 unique operating conditions (defined by axial load ($F_x$), radial load ($F_y$), and rotational speed), which are detailed in~\ref{sec:app_bearing_conds}.
Within each operating condition, the initial 50\% of the sequential data was allocated to the training and validation sets, with a random 80/20 split. 
The remaining 50\% of the sequential data formed the test set, representing the period when the virtual sensor would be actively predicting bearing load. This temporal split mirrors the real-world scenario where a physical sensor might have a limited lifespan due to factors such as battery drain or wear and tear. The
virtual sensor is trained and validated on data collected during the physical sensor's operational life, and then used to predict future load conditions.

\subsubsection{Heterogeneous Bearing Graph Construction}
\begin{figure}[tbhp]
  \centering
    \includegraphics[width=.9\linewidth]{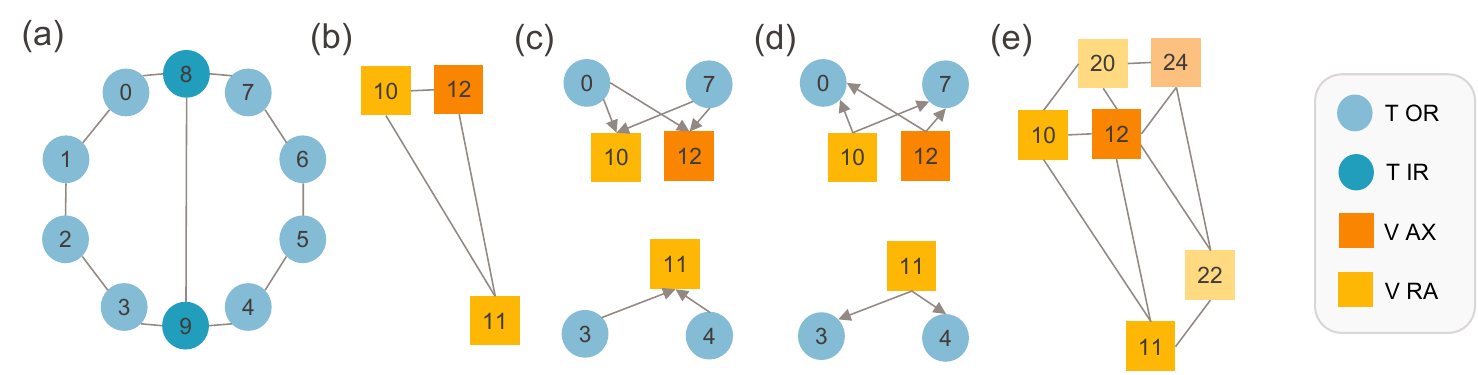}
  \caption{Heterogeneous graphs for bearing sensor network relationship modeling. (a) Temperature-Temperature (b) Vibration-Vibration (c) Temperature-Vibration (d) Vibration-Temperature (e) Connectivity across two test rig bearings (the connectivity between T nodes omitted for simplicity).}
  \label{fig:bearing_homogenous_graphs}
\end{figure}
We construct a heterogeneous graph with nodes representing sensors (temperature (T) and vibration (V)). Temperature nodes are further classified into inner ring (T IR) or outer ring (T OR) nodes. V nodes, which are installed on the outer ring, are distinguished by their load direction: radial (V RA) or axial (V AX). We model four types of relationships: T-T, V-V, T-V, and V-T. Here, T-T and V-V represent intra-modality interactions, while V-T and T-V represent inter-modality interactions. Node positions reflect physical sensor placement. Fig.~\ref{fig:bearing_homogenous_graphs}{(a)} and {(b)} illustrate the connectivity within a single bearing based on physical proximity. Additionally, IR nodes are connected due to relatively uniform temperatures across the inner ring. Given that the test rig consists of two bearings, we connect them based on proximity, as illustrated in Fig.~\ref{fig:bearing_homogenous_graphs}{(e)} for V nodes. We assume symmetrical (undirected) relationships within the same sensor type and model heterogeneous T-V and V-T relationships with directed edges, as demonstrated in Fig.~\ref{fig:bearing_homogenous_graphs}{(c)} and {(d)}.

\subsection{Case Study 2: Bridge}
\label{sec:case_study_bridge}
In this case study, we evaluate our proposed HTGNN model for bridge health monitoring, specifically focusing on estimating live load on the bridge deck from displacement and acceleration signals. Live load refers to the dynamic forces exerted on the bridge due to traffic, such as the weight of vehicles passing over it. Accurate live load estimation is essential for assessing structural integrity, identifying potential overloads or fatigue damage, and making informed decisions about maintenance and repair~\cite{liu2009bridge}. This capability is essential for ensuring the safety and longevity of bridge structures.

\subsubsection{Train-Track-Bridge Simulator and Data Acquisition}
Real-world bridge health monitoring data that captures the combined effects of train loads, speeds, and temperature variations is often scarce and not publicly available, especially data that includes both displacement and acceleration measurements. 
To address this, we leveraged the Train-Track-Bridge (TTB) simulator \cite{cantero2022ttb}, a recent finite element method-based tool offering highly realistic simulations of train, track, and bridge interactions under varying conditions, including the influence of track irregularities, a significant source of excitation for bridges.

Following the approach in Sarwar \textit{et al.}~\cite{sarwar2024probabilistic}, we simulated an ICE3 Velaro train with eight wagons on a bridge with a length of $L=50\si{m}$, a second moment of area $I = 51.3 \si{m^4}$, a mass per unit length $\rho = 69000 \si{kg/m}$, and a modulus of elasticity $E = 3.5 \times 10^{10} \si{N/m}$. To simulate realistic temperature fluctuations, hourly temperature data (0.1°C precision) from the Zurich Fluntern weather station was incorporated. This data was sourced from \href{https://www.visualcrossing.com/}{Visual Crossing}.
To estimate train loads, we utilized SBB passenger traffic statistics \cite{sbb_passenger_data}, which detail the percentage of long-distance trains at Zurich's main station per hour for 2023, differentiated by weekdays and weekends. Daily load factors (standard deviation 0.1) were randomly drawn from this distribution and applied to a base load of 42100 kg. Wagon-to-wagon load variations were modeled using a normal distribution (standard deviation 500 kg) around this final load, resulting in loads ranging from 42100 kg to 53500 kg.
Nine train runs were simulated per day, starting at 6:00 AM with two-hour intervals until 10:00 PM. To capture representative load patterns and comprehensive temperature variations, we simulated the first week of each month over an entire year, resulting in 756 unique train runs with diverse load, speed, and temperature conditions.

\subsubsection{Data Preprocessing}
Raw simulation data was first cropped to focus on the period when the train was fully on the bridge based on the magnitude of the displacement sensors, excluding the entry and exit phases. To simulate realistic sensor noise, additive white Gaussian noise with a signal-to-noise ratio (SNR) of 35 dB was introduced. The data, originally sampled at 1000 Hz, was then downsampled to 100 Hz using an 8th-order Chebyshev Type I filter. Finally, the dataset was divided into windows with a length of 60 samples (0.6 seconds at 100 Hz) and a stride of 5. We construct a heterogeneous graph for the bridge sensor network, following the same principles established for the bearing graph, see details in~\ref{sec:app_bridge_setup}.

\subsection{Train-Validation-Test Split}
To emulate the real-world deployment of a virtual sensor, the preprocessed dataset (490,398 samples) was split temporally by day. 
All odd-numbered days (48 days total) were allocated to the training and validation sets, with an 80/20 split, simulating the data collection phase under various weather and load conditions. Even-numbered days (36 days total) were used to form the test set, representing the period during which the virtual sensor would actively predict future loads based on the trained model.


\section{Evaluation Setup}
\label{sec:exp_design}
This section details the experimental design employed to evaluate the effectiveness of HTGNN in load prediction within a heterogeneous sensor network setting. Specifically, we introduce the baseline methods and their configurations in Sec.~\ref{sec:doe_baselines}, evaluation metrics in Sec.~\ref{sec:doe_evaluation}, training setups in Sec.~\ref{sec:doe_training} and experiment setups, including hardware specifications,  in Sec.~\ref{sec:doe_hardware}.

\subsection{Baseline Methods.} 
\label{sec:doe_baselines}
To provide a comprehensive evaluation of our proposed method, we benchmark its performance against several established approaches for virtual sensing~\cite{sun2021survey}, each selected for its relevance to the problem domain.
\revtext{We chose Convolutional Neural Networks (CNNs) and Recurrent Neural Networks (RNNs) as our primary baselines due to their proven effectiveness in handling temporal dynamics~\cite{sun2021survey}, a key aspect of our virtual sensing task with heterogeneous temporal dynamics. CNNs are well suited for local dynamics modeling and frequency domain processing, while RNNs are capable of capturing long-term dependencies and overall dynamics modeling. Both these capabilities are crucial for effectively capturing the complex temporal relationships in our problem. Specifically, we}
compare against the Bidirectional LSTM (BiLSTM) due to its widespread use in sequence modeling, and the 1D Convolutional Neural Network (1D-CNN) for its effectiveness in signal processing tasks. The 1D Gated CNN (1D-GCNN) is included to assess the impact of gating mechanisms similar to those in our approach. 
\revtext{To assess the benefits of multiscale modeling, we include TimeMixer~\cite{wang2023timemixer}, originally designed for forecasting tasks. We adapt its architecture for our regression task by removing the decoder component and introducing a regression head, following the implementation provided in~\cite{wang2024deep} adapted for classification tasks.}
\revtext{
Finally, we include the Multivariate Time-series Graph Attention Network (MTGAT)
, originally introduced by \cite{zhao2020multivariate} as an anomaly detection framework and later adapted by \cite{niresi2023spatial} for the sensor calibration task,} as a state-of-the-art homogeneous \revtext{temporal graph regression model. This serves as} a benchmark for the effectiveness of our heterogeneous graph-based approach.
For each model, we added a head that converts from the final hidden dimension of the model to the output, implemented as a three-layered MLP. All models utilized the SiLU (Sigmoid Linear Unit) activation function. 
For further details on the hyperparameter tuning of the baseline methods and our proposed HTGNN model, please refer to \ref{sec:app_hyper}.

\subsection{Evaluation Metrics}
\label{sec:doe_evaluation}
The selection of evaluation metrics was tailored to the characteristics of each dataset. For the bearing dataset, which does not contain zero values, we employed Mean Absolute Percentage Error (MAPE) to assess model performance. MAPE allows for direct comparison of the axial and radial load predictions due to its scale invariance. Conversely, the bridge dataset may contain values near zero. In this case, we utilized Normalized Root Mean Squared Error (NRMSE) to provide a more robust evaluation, as MAPE can be biased when predictions are close to zero.

\subsection{Training Setup} 
\label{sec:doe_training}
All models were trained using the AdamW optimizer \cite{loshchilov2017decoupled} with an initial learning rate of $5 \times 10^{-3}$, minimizing the mean squared error (L2 loss). 
Training proceeded for a maximum of 150 epochs. To prevent overfitting, we employed early stopping with patience of 20 epochs based on the validation loss, ensuring a minimum of 50 training epochs. Additionally, ReduceLROnPlateau scheduling was implemented, which reduced the learning rate by a factor of 0.9 if the validation loss did not improve for 10 consecutive epochs. To further enhance convergence, a learning rate warm-up strategy was applied, gradually decreasing the learning rate from the initial value to a minimum of $1 \times 10^{-4}$ within the first few iterations (200 for bearing datasets, 500 for bridge datasets).

\subsection{Experimental Setup}
\label{sec:doe_hardware}
The implementation of all methods, including the proposed method and baselines, was carried out using PyTorch 1.12.1~\cite{paszke2019pytorch} with CUDA 12.0 and the PyTorch Geometric 2.2.0~\cite{fey2019fast}. 
Computations for both datasets were executed on a GPU cluster equipped with NVIDIA A100 80GB GPUs. We employed neptune.ai for experiment tracking and management.

\section{Results}
\label{sec:result}

\subsection{Bearing Load Prediction}
\subsubsection{Overall Quantitative Performance}
In the bearing load prediction task, we evaluated the performance of our proposed HTGNN model against several baseline methods using two key metrics: Normalized Root Mean Squared Error (NRMSE) and Mean Absolute Percentage Error (MAPE). The results, presented in Tab.~\ref{tab:bearing_result_nrmse} and Tab.~\ref{tab:bearing_result_mape}, show that model performance varies across different rotational speeds and load types. In the following, we present a detailed analysis of these results that reveals several key insights into the strengths and weaknesses of the baseline methods, their performance compared to the proposed HTGNN model, as well as the factors influencing their performance.

\begin{table}[tbh]
\centering
\caption{Model performance (NRMSE) for bearing load predictions across different rotational categories. Values in ± indicate the 95\% confidence interval over five runs.}
\label{tab:bearing_result_nrmse}
\renewcommand{\arraystretch}{1.1} 
\resizebox{.9\columnwidth}{!}{%
\begin{tabular}{lcccccc}
\toprule
\multirow{2}{*}{\textbf{Model}} & \multicolumn{5}{c}{\textbf{Rotational Speed [r/min]}} & \multirow{2}{*}{\textbf{Avg.}} \\
\cmidrule(lr){2-6}
        & \textbf{10} & \textbf{20} & \textbf{30} & \textbf{40} & \textbf{50} & \\
\midrule
  \textbf{BiLSTM}  &  {0.021} \scriptsize{± 0.006} &  {0.020} \scriptsize{± 0.008} &  {0.019} \scriptsize{± 0.008} &  {0.021} \scriptsize{± 0.008} &  {0.015} \scriptsize{± 0.003} &  {0.019} \\
  \textbf{1D-CNN}   &  0.038 \scriptsize{± 0.008} &  0.037 \scriptsize{± 0.012} &  0.038 \scriptsize{± 0.011} &  0.025 \scriptsize{± 0.006} &  0.022 \scriptsize{± 0.004} &  0.032 \\
  \textbf{1D-GCNN}  &  0.023 \scriptsize{± 0.005} &  0.029 \scriptsize{± 0.018} &  0.024 \scriptsize{± 0.008} &  {0.017} \scriptsize{± 0.005} &  {0.022} \scriptsize{± 0.005} &  0.023  \\
  \revtext{\textbf{TimeMixer}} &  \revtext{\underline{0.019} \scriptsize{± 0.007}} & \revtext{\underline{0.010} \scriptsize{± 0.003}} &  \revtext{\underline{0.013} \scriptsize{± 0.008}} & \revtext{\underline{0.011} \scriptsize{± 0.003}} &  \revtext{\textbf{0.005} \scriptsize{± 0.001}} &  \revtext{\underline{0.012}}\\
  \textbf{MTGAT} &  0.024 \scriptsize{± 0.007} &  0.025 \scriptsize{± 0.009} &  {0.020} \scriptsize{± 0.008} &  0.021 \scriptsize{± 0.009} &  {0.015} \scriptsize{± 0.003} &  0.021  \\
  \textbf{HTGNN} &  \textbf{0.008} \scriptsize{± 0.004} &  \textbf{0.004} \scriptsize{± 0.001} &  \textbf{0.010} \scriptsize{± 0.010} &  \textbf{0.005} \scriptsize{± 0.003} &  \underline{0.006} \scriptsize{± 0.002} &  \textbf{0.007} \\
\bottomrule
\end{tabular}%
}
\end{table}

\textbf{Baseline method performance.} 
The complex, interdependent interactions between temperature and vibration measurements in bearing load prediction pose a challenge for many models. These interactions vary not only with load but also with rotational speed. While higher loads generally decrease vibration magnitudes \cite{bal2022effect} but increase temperature change rates \cite{mohammed2019situ}, the influence of load on these change rates can be altered by the rotational speed. At higher speeds, both temperature and vibration magnitudes often exhibit steeper increases or decreases in response to load changes.
Comparison models, such as MTGAT, which assume homogeneous relations between sensor nodes, struggle to capture these heterogeneous dynamics. Similarly, sequence-based models like 1D-CNN and BiLSTM, primarily focusing on temporal dependencies within a single sensor modality, also face limitations in modeling the complex relationships between temperature and vibration across varying rotational speeds. \revtext{
TimeMixer demonstrates greater stability in these worst-case scenarios, with the highest MAPE for both axial and radial load prediction remaining below 25\%. This stability highlights the benefits of multiscale modeling in improving model robustness. However, while TimeMixer can capture some multiscale dynamics, it does not explicitly capture inter-scale interactions or fully account for the impact of exogenous variables like rotational speed, which limits its performance under rare operating conditions and leads to worse predictions at lower speeds.}

\begin{table}[bthp]
\centering
\renewcommand{\arraystretch}{1.1} 
\caption{
Model performance (MAPE) for bearing load predictions ($F_x$ and $F_y$) across different rotational speeds. Values annotated by $<$ indicate the 95\% maximum MAPE observed across five runs.}
\label{tab:bearing_result_mape}
\resizebox{.9\columnwidth}{!}{%
\begin{tabular}{lllllllr}
\toprule
& \multirow{2}{*}{\textbf{Model}} & \multicolumn{5}{c}{\textbf{Rotational Speed [r/min]}} & \multirow{2}{*}{\textbf{Avg.}} \\
\cmidrule(lr){3-7}
        &  & \textbf{10} & \textbf{20} & \textbf{30} & \textbf{40} & \textbf{50} & \\
\midrule
\multirow{5}{*}{\textbf{MAPE\textsubscript{\(F_x\)} (\%)}} 
        & \textbf{BiLSTM} & {2.6} \scriptsize{$<$ 7.3} &  4.1 \scriptsize{$<$ 19.3} &  3.1 \scriptsize{$<$ 11.1} &   {3.5} \scriptsize{$<$ 11.5} &  {2.7} \scriptsize{$<$ 8.1} &  {3.2} \\
        & \textbf{1D-CNN} &  5.4 \scriptsize{$<$ 13.3} &  7.1 \scriptsize{$<$ 28.4} &  7.4 \scriptsize{$<$ 29.8} &  4.4 \scriptsize{$<$ 12.2} &  5.2 \scriptsize{$<$ 16.8} &  5.9 \\
        & \textbf{1D-GCNN} & 3.4 \scriptsize{$<$ 10.3} &  \underline{2.7} \scriptsize{$<$ 11.0} &  4.9 \scriptsize{$<$ 22.3} &  {3.2} \scriptsize{$<$ 14.4} &  6.4 \scriptsize{$<$ 28.6} &  4.1 \\       
        & \revtext{\textbf{TimeMixer}} &  \revtext{{2.0} \scriptsize{$<$ 8.6}} & \revtext{\underline{2.7} \scriptsize{$<$ 7.9}} &  \revtext{\underline{2.2} \scriptsize{$<$ 8.9}} & \revtext{\underline{1.9} \scriptsize{$<$ 6.3}} &  \revtext{\underline{1.2} \scriptsize{$<$ 4.2}} &  \revtext{\underline{2.0}}\\        
        & \textbf{MTGAT} &   2.9 \scriptsize{$<$ 9.3} &  6.9 \scriptsize{$<$ 35.5} &   {3.0} \scriptsize{$<$ 9.1} &  4.2 \scriptsize{$<$ 20.9} &  3.4 \scriptsize{$<$ 10.6} &  4.1 \\
        & \textbf{HTGNN} &  \textbf{0.7} \scriptsize{$<$ 2.5} &   \textbf{0.4} \scriptsize{$<$ 1.3} &   \textbf{1.3} \scriptsize{$<$ 8.8} &   \textbf{0.8} \scriptsize{$<$ 3.0} &  \textbf{0.5} \scriptsize{$<$ 1.8} &  \textbf{1.0} \\
\midrule
\multirow{5}{*}{\textbf{MAPE\textsubscript{\(F_y\)} (\%)}} 
        & \textbf{BiLSTM} & {4.5} \scriptsize{$<$ 19.2} &   1.9 \scriptsize{$<$ 5.7} &   1.9 \scriptsize{$<$ 4.8} &   2.3 \scriptsize{$<$ 5.3} &  5.4 \scriptsize{$<$ 16.8} &  3.2 \\
        & \textbf{1D-CNN} &  5.1 \scriptsize{$<$ 20.8} &  3.8 \scriptsize{$<$ 15.7} &   2.7 \scriptsize{$<$ 5.3} &   3.0 \scriptsize{$<$ 9.3} &  5.2 \scriptsize{$<$ 12.3} &  4.0 \\
        & \textbf{1D-GCNN} &  5.1 \scriptsize{$<$ 19.6} &  8.1 \scriptsize{$<$ 12.2} &  3.7 \scriptsize{$<$ 16.5} &  3.0 \scriptsize{$<$ 11.7} &  6.4 \scriptsize{$<$ 19.5} &  5.2 \\
        & \revtext{\textbf{TimeMixer}} &  \revtext{{4.6} \scriptsize{$<$ 21.8}} & \revtext{\underline{1.4} \scriptsize{$<$ 8.6}} &  \revtext{\underline{1.1} \scriptsize{$<$ 3.7}} & \revtext{\underline{1.3} \scriptsize{$<$ 3.7}} &  \revtext{\textbf{1.5} \scriptsize{$<$ 4.3}} &  \revtext{\underline{2.0}}\\
        & \textbf{MTGAT} &  \underline{4.3} \scriptsize{$<$ 18.6} &   {1.6} \scriptsize{$<$ 4.9} &  {1.4} \scriptsize{$<$ 4.4} &   {1.9} \scriptsize{$<$ 6.4} &  {3.8} \scriptsize{$<$ 10.7} &  {2.6} \\
        & \textbf{HTGNN} &   \textbf{2.2} \scriptsize{$<$ 14.2} &   \textbf{1.0} \scriptsize{$<$ 7.6} &   \textbf{0.4} \scriptsize{$<$ 1.1} &   \textbf{0.6} \scriptsize{$<$ 1.9} &   \underline{2.3} \scriptsize{$<$ 11.0} &  \textbf{1.3} \\
\bottomrule
\end{tabular}%
}
\end{table}

\begin{figure}[tbhp]
  \centering
    \includegraphics[width=.8\linewidth]{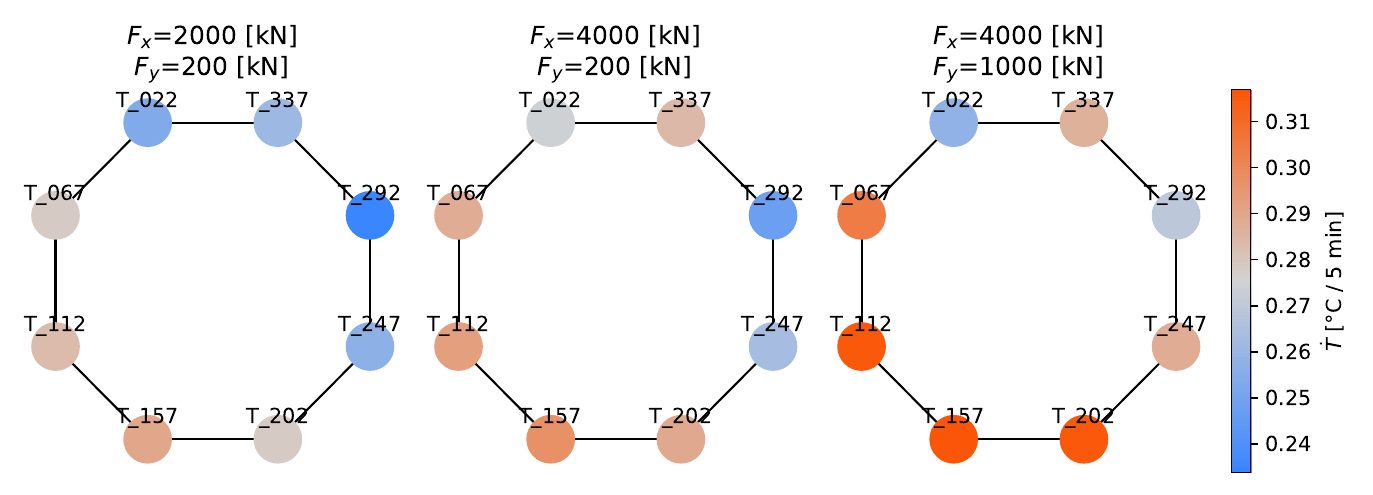}
  \caption{Temperature change rates at different bearing locations under varying axial ($F_x$) and radial ($F_y$) load conditions and a constant rotational speed ($30 [r/min]$).}
  \label{fig:temp_diff_distribution}
\end{figure}

\textbf{Benefits of heterogeneous interaction modeling.} 
In contrast, HTGNN's architecture explicitly models these heterogeneous relationships and their dependence on operating conditions. This enables HTGNN to effectively extract information from both temperature and vibration sensors across varying rotational speeds, resulting in improved load prediction.
By capturing these complex dependencies, HTGNN outperforms \revtext{most baseline} models across all rotational speeds and load types, as evidenced by the lowest NRMSE values with tight confidence intervals (Tab.~\ref{tab:bearing_result_nrmse}) and the lowest maximum MAPE values (Tab.~\ref{tab:bearing_result_mape}). The low variance in HTGNN's performance metrics further underscores its robustness and reliability, making it a promising solution for virtual load sensing in complex industrial systems.

\textbf{Graph as physical prior for radial load distribution.} 
As shown in Tab.~\ref{tab:bearing_result_mape}, graph-based models like HTGNN and MTGAT exhibit superior performance in radial load prediction, evident by their lower MAPE values compared to other models. This advantage stems from their ability to leverage the bearing's physical connectivity as an inductive bias within the graph structure. 
Temperature signals not only reflect load magnitude through an increased rate of change under higher loads~\cite{mohammed2019situ} but also reveal the spatial distribution of the load through localized temperature increases in the loaded zones~\cite{sun2024simulation}. 
This relationship is illustrated in Fig.~\ref{fig:temp_diff_distribution}, where higher temperature changes are concentrated in the specific areas experiencing radial load, while both axial and radial loads contribute to the overall magnitude of temperature increase.
By incorporating the physical connectivity of temperature sensors within the bearing system, HTGNN effectively leverages this spatial prior knowledge to learn the relationship between temperature distribution and radial load, leading to improved prediction performance.

\textbf{Effectiveness of the gating mechanism.} 
Incorporating a gating mechanism for rotational speed embedding significantly improves performance, as evidenced by 1D-GCNN's reduced NRMSE and axial MAPE compared to 1D-CNN across all speeds. This suggests that the gating mechanism allows the model to better capture the influence of rotational speed on vibration signals, which are particularly informative for axial load prediction. However, while 1D-GCNN outperforms 1D-CNN in terms of NRMSE, its radial MAPE remains higher, likely due to its inability to explicitly model the spatial relationships between temperature sensors crucial for understanding radial load distribution.
HTGNN, which also employs a gating mechanism, consistently achieves the best performance for axial MAPE across all rotational speeds. This is likely due to its ability to selectively activate the most informative frequency components within the vibration signals based on the specific operating conditions. By effectively leveraging both the spatial relationships between temperature sensors and the frequency components of vibration signals, HTGNN  demonstrates the efficacy of gating mechanisms in enhancing load prediction. This ability to dynamically adapt to different operating conditions results in superior load prediction performance, showcasing HTGNN's robustness and versatility.

\subsubsection{In-Depth Analysis of Load Predictions}
\begin{figure}[tbh]
  \centering
    \includegraphics[width=.65\linewidth]{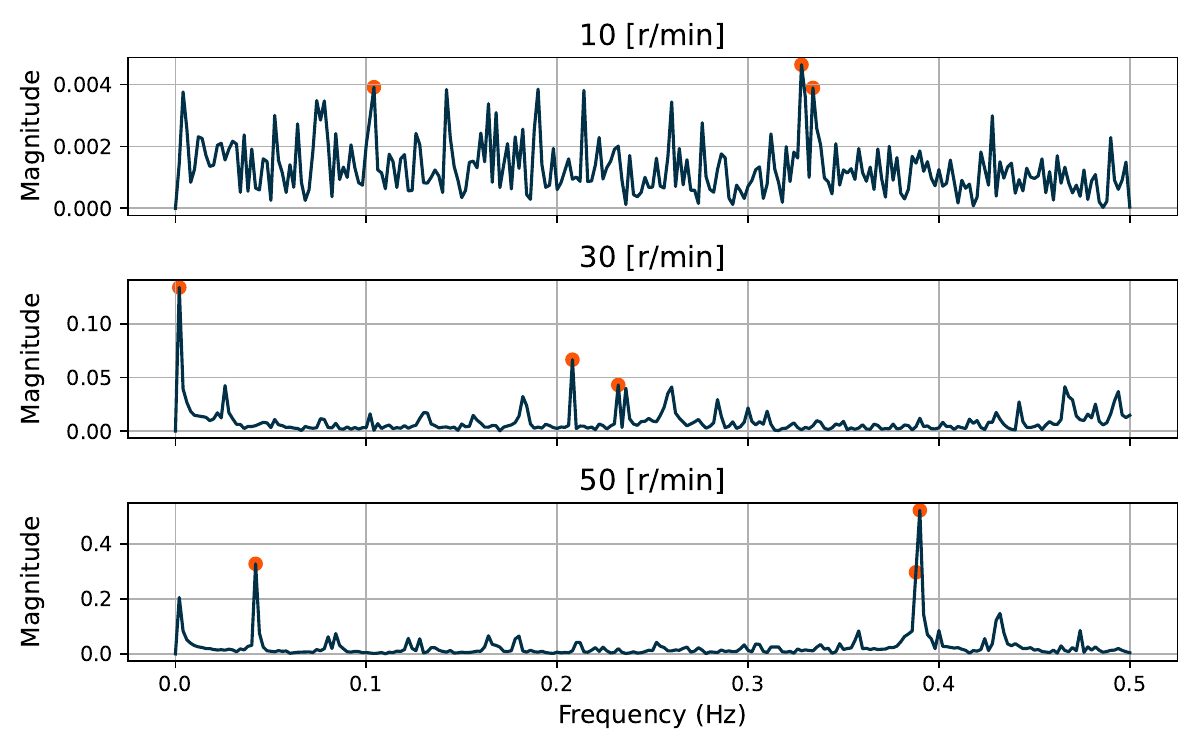}
\caption{Frequency spectra of bearing radial vibration signals at different rotational speeds (10, 30, and 50 [r/min]) under a constant load of $F_x$=2000 [kN] and $F_y$ = 200 [kN]. The top three dominant frequencies in each spectrum are highlighted with red dots.}
  \label{fig:bearing_fft}
\end{figure}
\begin{figure}[tbh]
  \centering
    \includegraphics[width=1\linewidth]{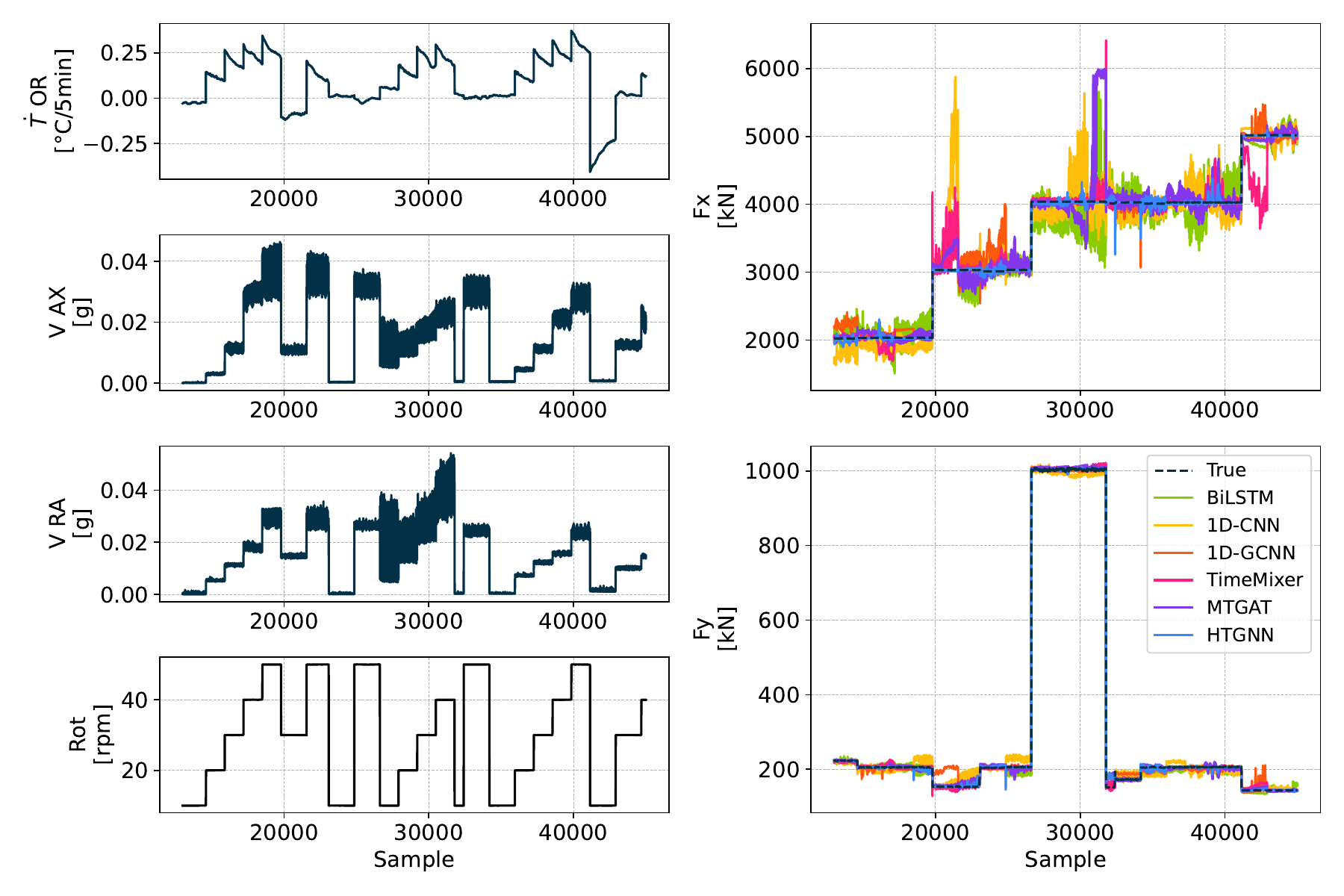}
  \caption{Examples of input signals and load prediction performance. 
  The left four plots illustrate a sensor signal for each of the following: the rate of temperature change in the outer ring of bearing TB1 ($\dot{T}$ OR), axial vibration (V AX), radial vibration (V RA), and rotational speed (Rot). The right two plots display the true and predicted values of the bearing forces in the axial ($F_x$) and radial ($F_y$) directions for the best-performing instance of each model.}
  \label{fig:bearing_zoomin_plot}
\end{figure}

\textbf{Impact of rotational speed on signal characteristics.} Fig.~\ref{fig:bearing_zoomin_plot} showcases examples of input sensor signals and the corresponding axial and radial load predictions for the best-performing instance of each model. 
The left four plots illustrate examples of each type of input time series: temperature change rate in the outer ring (T OR), axial vibration (V AX), radial vibration (V RA), and rotational speed (Rot).
As shown, the intensity of both axial and radial vibrations, as well as the rate of temperature change, increase with rotational speed.
This relationship between rotational speed and vibration signals is further explored in Fig.~\ref{fig:bearing_fft}, which presents the frequency spectra of the radial vibration signals at different rotational speeds under constant load. Although these spectra are derived from aggregated signals, they reveal a general trend of frequency content shifting towards higher frequencies with increasing rotational speed, with new, higher-frequency components emerging. 
This observation highlights the importance of accurately modeling the relationship between rotational speed and the measurement signals for precise load prediction. 

\textbf{Challenges in load prediction.}
The right two plots of Fig.~\ref{fig:bearing_zoomin_plot} illustrate the varying degrees of deviation from true load values observed in the performance of baseline models, particularly for axial loads. 
The relatively flat frequency spectra observed at low speeds (Figure~\ref{fig:bearing_fft}) indicate a lower signal-to-noise ratio, which may hinder the performance of frequency-based models like 1D-CNN and 1D-GCNN that rely on prominent frequency-domain features. While 1D-GCNN's gating mechanism offers some mitigation compared to 1D-CNN, both models remain susceptible to significant deviations under these conditions.
MTGAT, a homogeneous GNN, also exhibits occasional extreme errors. This might be due to its aggregation of inherently heterogeneous features (e.g., temperature and vibration), potentially hindering its ability to extract meaningful representations.
Additionally, BiLSTM, as a sequential model, may face challenges in extracting meaningful frequency information from vibration signals, particularly when signal trends are less apparent.
\revtext{TimeMixer, while generally performing well, exhibits sensitivity to changing operating conditions (i.e. during shifts in rotational speed). This sensitivity likely stems from its inability to explicitly account for the impact of exogenous variables on the measured signals.}

\begin{figure}[tbh]
  \centering
    \includegraphics[width=1\linewidth]{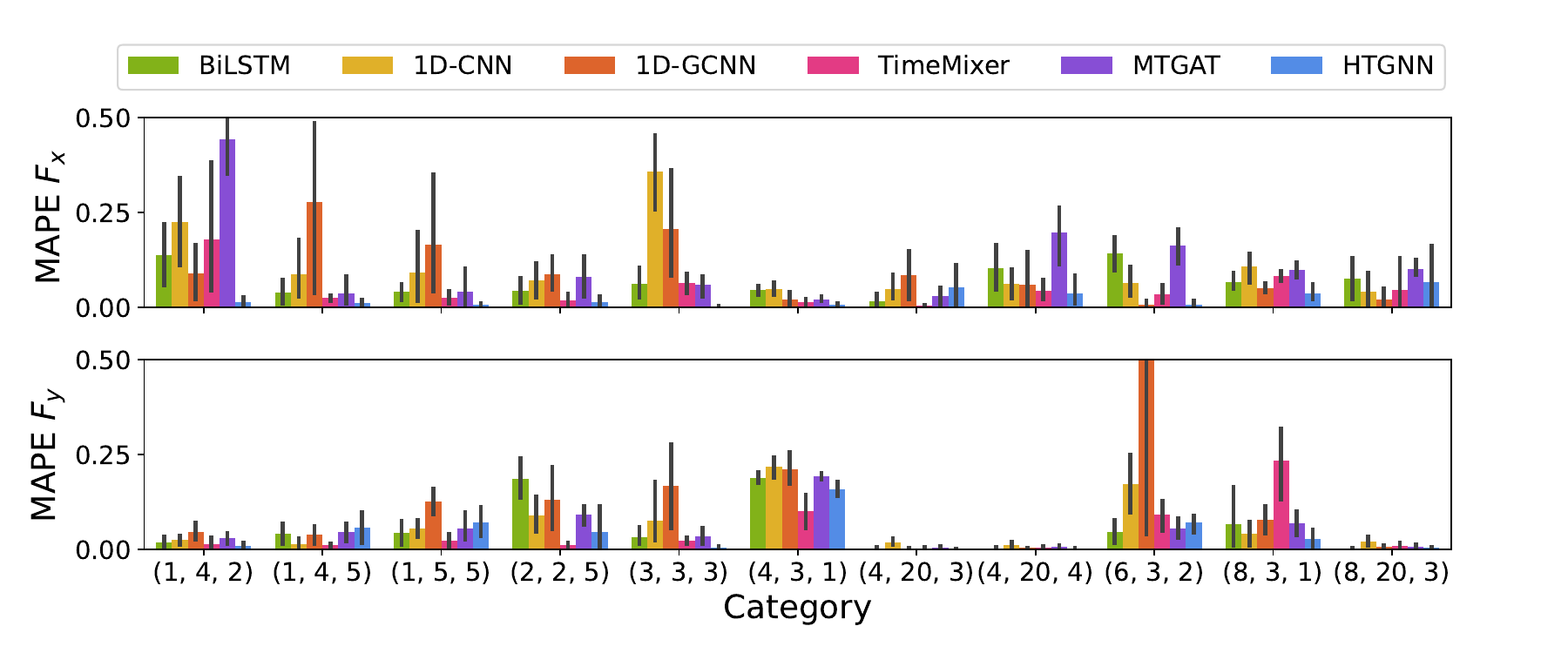}
  \caption{Average test set performance (MAPE) of each model's worst two scenarios across five runs. Error bars show the 95\% confidence interval of the mean. The scenarios,  representing combinations of bearing load and rotational speed, are defined as the $F_x (\times 1000)$ [kN], $F_y (\times 50)$ [kN], and rotational speed ($\times 10$) [r/min].}
  \label{fig:bearing_mape_worst}
\end{figure}

\textbf{Worst case performance assessment.}
To further assess the robustness of each model, we examine their worst-case performance under challenging operating conditions. Fig.~\ref{fig:bearing_mape_worst} presents the average MAPE for their two most challenging scenarios across five runs. The error bars represent the 95\% confidence intervals. Notably, these categories often involve low axial and radial loads, which inherently lead to higher MAPE values due to the nature of this metric (i.e., dividing by a small actual value). 
For axial load prediction, 1D-CNN and 1D-GCNN consistently struggle, particularly under low axial loads and high rotational speeds. These conditions often lead to significant fluctuations and variations in vibration signal magnitudes.  CNN-based methods may not capture effectively due to their primary focus on the amplitude of frequency components rather than the overall signal trends and changes, which are crucial for accurate load prediction.  While 1D-GCNN's gating mechanism offers some improvement over 1D-CNN by modulating extracted features based on rotational speed, it does not consistently rectify this limitation across all operating conditions.
Similarly, BiLSTM exhibits high axial MAPE under both high radial loads with high speeds and low radial loads with low speeds. 
These conditions lead to extreme vibration amplitudes (high or low), which BiLSTM, as a sequential model, may struggle to handle effectively due to the complexities in extracting relevant features from such signals.
These findings highlight the complex nonlinear relationship between vibration, temperature, rotational speed, and load, posing a particular challenge for baseline models that lack the ability to effectively capture both high-frequency and low-frequency features simultaneously.
\revtext{TimeMixer demonstrates greater stability in these worst-case scenarios, with the highest MAPE for both axial and radial load prediction remaining below 25\%. This stability highlights the benefits of multiscale modeling in improving model robustness. However, TimeMixer, like the other baselines, does not explicitly account for the influence of exogenous variables on the measured signals and therefore performs worse under rarely represented operating conditions, such as low load and low rotational speed.}

\textbf{HTGNN's robust virtual sensing across operating conditions}.
In contrast to the limitations exhibited by baseline models, particularly under challenging operating conditions, HTGNN consistently demonstrates superior accuracy across all rotational speeds and load types (Fig.~\ref{fig:bearing_zoomin_plot}, right two plots). 
HTGNN's ability to closely track the true load values, even as signal characteristics change with rotational speed, demonstrates its effectiveness in capturing the underlying dynamics of the bearing system. 
Furthermore, HTGNN demonstrates a strong consistency and robustness, maintaining relatively low MAPE values across most scenarios, even under the most challenging and under-represented operating conditions (Fig.~\ref{fig:bearing_mape_worst}). 
This robust performance establishes HTGNN as a promising solution for reliable virtual load sensing, enabling effective bearing monitoring and maintenance.

\subsection{Bridge Load Prediction}
\begin{table}[bth]
\centering
\caption{Model performance (NRMSE) for train load prediction across temperature categories (mean and 95\% confidence interval)}
\label{tab:result_bridge_nrmse} 
\renewcommand{\arraystretch}{1.1} 
\resizebox{.8\columnwidth}{!}{%
    \begin{tabularx}{1.08\linewidth}{l *{6}{c}} 
    \toprule
        & \multicolumn{4}{c}{\textbf{Temperature [°C]}} & \multicolumn{1}{c}{\multirow{2}{*}{\textbf{Avg.}}} \\
    \cmidrule(lr){2-5}
    \textbf{Model} & \textbf{$<$ 0} & \textbf{0-10} & \textbf{10-20} & \textbf{$>$ 20} & \\
    \midrule
    \textbf{BiLSTM}   & 0.089 \scriptsize{$\pm$ 0.021} & \underline{0.054} \scriptsize{$\pm$ 0.003} & 0.046 \scriptsize{$\pm$ 0.003} & 0.049 \scriptsize{$\pm$ 0.004} & 0.059\\
    \textbf{1D-CNN} & 0.125 \scriptsize{$\pm$ 0.028} & 0.066 \scriptsize{$\pm$ 0.004} & \textbf{0.043} \scriptsize{$\pm$ 0.003} & 0.049 \scriptsize{$\pm$ 0.004} & 0.070\\
    \textbf{1D-GCNN}  & \underline{0.080} \scriptsize{$\pm$ 0.019} & \underline{0.054} \scriptsize{$\pm$ 0.003} & 0.045 \scriptsize{$\pm$ 0.003} & \textbf{0.046} \scriptsize{$\pm$ 0.004} & \underline{0.056} \\
    \revtext{\textbf{TimeMixer}}  & \revtext{{0.094} \scriptsize{{$\pm$ 0.067}}} & \revtext{ 0.073 \scriptsize{$\pm$ 0.007}} & \revtext{0.050 \scriptsize{$\pm$ 0.040}} & \revtext{0.055 \scriptsize{$\pm$ 0.041}} & \revtext{0.068}  \\
    \textbf{MTGAT}  & 0.118 \scriptsize{$\pm$ 0.034} & 0.095 \scriptsize{$\pm$ 0.007} & 0.047 \scriptsize{$\pm$ 0.003} & 0.052 \scriptsize{$\pm$ 0.005} & 0.078  \\
    \textbf{HTGNN}  & \textbf{0.072} \scriptsize{$\pm$ 0.015} & \textbf{0.048} \scriptsize{$\pm$ 0.002} & \underline{0.044} \scriptsize{$\pm$ 0.003} & \underline{0.047} \scriptsize{$\pm$ 0.004} & \textbf{0.053} \\
    \bottomrule
    \end{tabularx}
}
\end{table}
\subsubsection{Overall Performance}
Tab.~\ref{tab:result_bridge_nrmse} presents the model performance in predicting train loads across different temperature categories using Normalized Root Mean Squared Error (NRMSE) with 95\% confidence intervals. 
Notably, all models exhibit a slight increase in NRMSE as temperature decreases. This could be attributed to the inherent difficulty of modeling bridge behavior at lower temperatures or to the fact that the training dataset was not sufficiently representative and did not contain enough observations of bridge behavior under such conditions.
HTGNN consistently demonstrates competitive performance across all temperature ranges, showing particular strength at lower temperatures where other models experience higher NRMSE.
1D-GCNN, with its gating mechanism, also exhibits strong performance, particularly at higher temperatures. 
1D-CNN and BiLSTM generally perform well, except at low temperatures. 
MTGAT \revtext{and TimeMixer} consistently \revtext{show} the highest NRMSE values. This underperformance, in contrast to \revtext{their} competitive results on the bearing dataset, may be attributed to the increased heterogeneity of the bridge sensor dataset. Unlike the bearing dataset, which is dominated by low-frequency temperature signals, the bridge dataset contains a balanced distribution of high-frequency (acceleration) and low-frequency (displacement) signals\revtext{, further compounded by the pronounced impact of the exogenous variable (temperature) on these signals.} The increased heterogeneity could pose challenges for MTGAT's homogeneous graph structure \revtext{as well as TimeMixer's limited ability to capture exogenous influences}, limiting \revtext{their} ability to fully capture the complex interactions between different sensor modalities.

\subsubsection{In-Depth Performance Evaluation}
\begin{figure}[tbh]
  \centering
    \includegraphics[width=.9\linewidth]{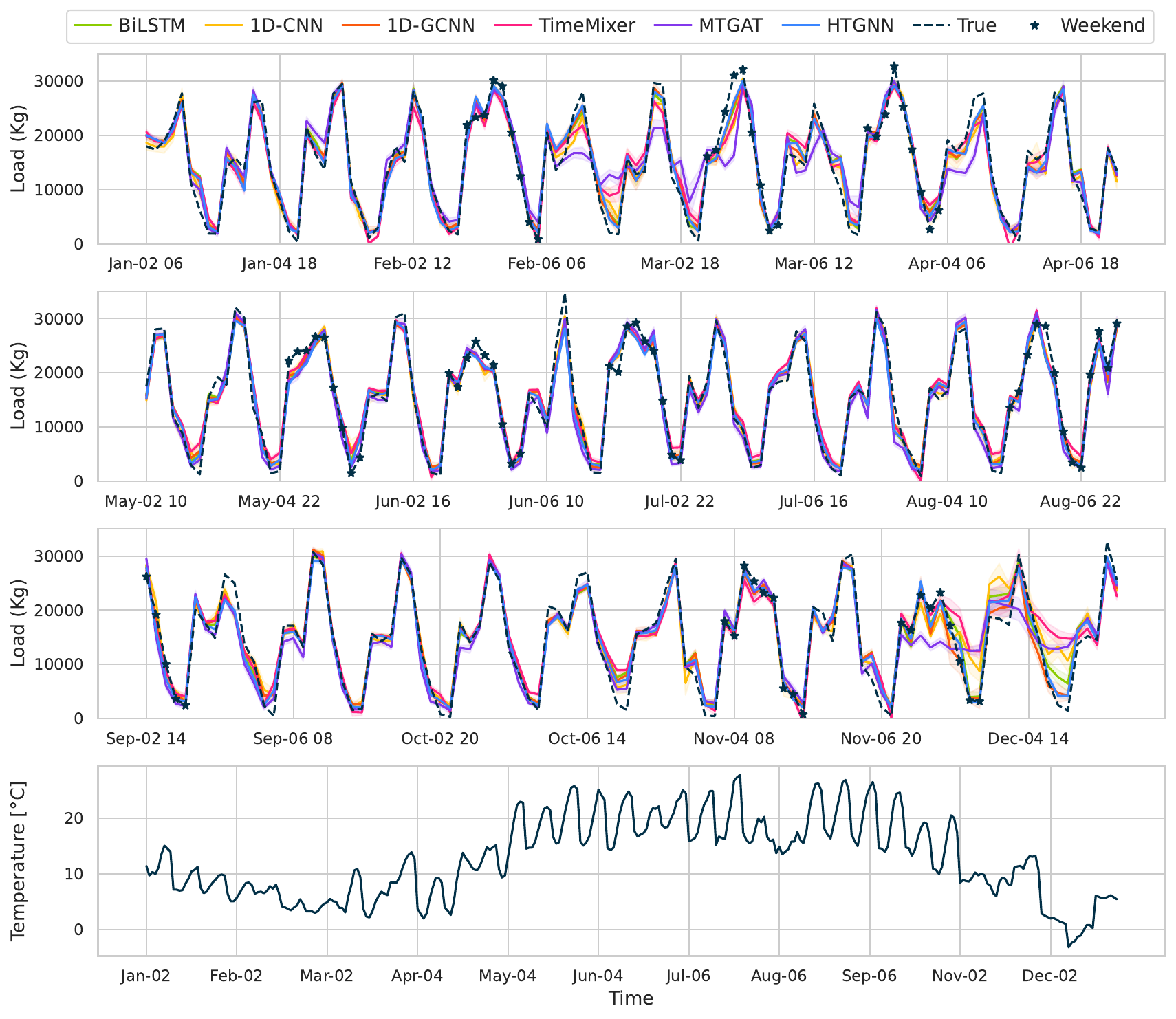}
  \caption{Train load prediction performance comparison for 2023. The areas around the predicted lines represent the 95\% confidence intervals over five runs. Weekends are marked with stars. The bottom plot shows the corresponding temperature over the same period.}
  \label{fig:bridge_result_by_case}
\end{figure}

\textbf{Model performance across varying conditions.} 
Fig.~\ref{fig:bridge_result_by_case} presents the train load prediction performance of all models for the year 2023, with 95\% confidence intervals over five runs shown as shaded areas. 
The bottom plot displays the corresponding temperature variations, while weekends are marked with stars. 
Notably, all models exhibit increased prediction error during colder months (e.g., March and December) and on weekends, as indicated by wider confidence intervals and larger deviations from the true load (black dashed line). This decrease in performance is particularly pronounced for the baseline models, which show significant fluctuations and inaccuracies during these periods. In contrast, HTGNN demonstrates superior performance across all conditions, consistently maintaining tight confidence intervals and closely tracking the true load. 

\begin{figure}[tbh]
    \centering
    \subfloat[Raw acceleration signals]{
            \includegraphics[width=0.5\linewidth]{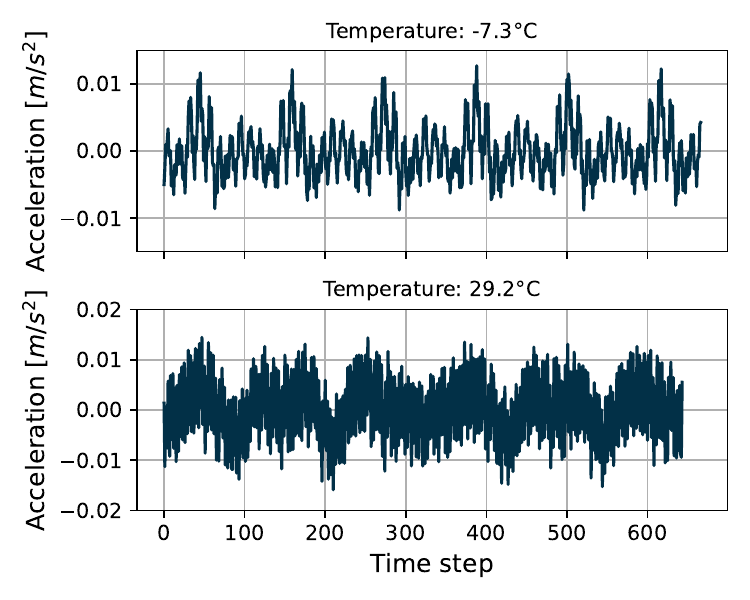}}
    \subfloat[Spectral centroid versus temperature.]{
        \includegraphics[width=.48\linewidth]{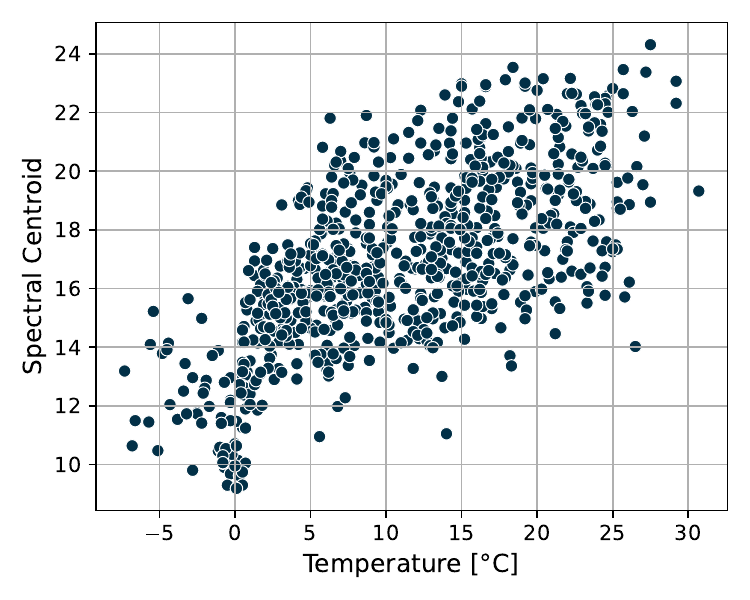}} 
    \caption{Impact of temperature on the frequency characteristics of the acceleration signal from accelerometer 0. }
    \label{fig:temp_impact_bridge}
\end{figure}

\textbf{Impact of temperature on signal characteristics}: 
To further investigate the influence of \revtext{the exogenous variable, temperature, }on bridge dynamics and its implications for load prediction, we analyze the frequency characteristics of acceleration signals under varying temperatures. 
Fig.~\ref{fig:temp_impact_bridge}(a) shows raw acceleration signals at -7.3°C and 29.2°C, revealing notable differences in signal amplitude and pattern. This temperature-dependent shift in frequency content is further analyzed in Fig.~\ref{fig:temp_impact_bridge}(b), where the spectral centroid (a measure of the dominant frequency) exhibits a positive correlation with temperature. 
This finding suggests that higher temperatures lead to an increase in the dominant frequency of acceleration signals. Such variations in signal characteristics could contribute to the observed decrease in prediction accuracy for all models at lower temperatures, likely due to the model's inherent limitations in capturing the complex non-linear relationship between temperature and the frequency response of the bridge structure.

\textbf{Effectiveness of the Gating Mechanism in 1D-GCNN}. 
During colder months, 1D-GCNN performs noticeably better than 1D-CNN, highlighting the effectiveness of the gating mechanism using temperature as a gating signal. This approach helps select frequency components more relevant to temperature conditions, demonstrating the benefits of incorporating contextual information into the model architecture. This contextual adaptation allows 1D-GCNN to maintain higher prediction accuracy by dynamically adjusting to the temperature-dependent variations in the acceleration signals.

\revtext{
\textbf{Limitation of simple multiscale modeling}. 
While TimeMixer's multiscale mixing strategy effectively captures temporal dependencies at different resolutions, it fails to explicitly model the heterogeneous interactions between these scales, particularly under strong exogenous influences.  This limitation is particularly evident in the bridge dataset, where temperature significantly affects both acceleration and displacement signals. TimeMixer's inability to capture these interactions and the exogenous influence on different scales likely contributes to its deteriorated performance at lower temperatures.}

\textbf{Limitation of homogenous graph models.} 
MTGAT's vulnerability to temperature variations and its generally inferior performance compared to other baseline models can be attributed to its homogeneous graph structure and its handling of temperature as a node within the graph. 
By assuming that all sensor nodes are of the same type and that relationships are homogenous, MTGAT may struggle to accurately represent the diverse and complex interactions between different sensor modalities in a heterogeneous sensor network. This homogenous assumption acts as a negative bias, hindering the model's ability to leverage the unique characteristics of each sensor type. Additionally, treating temperature as a node with the same relationships to displacement and acceleration signals might oversimplify the distinct ways in which these signals respond to temperature changes. This oversimplification can lead to overfitting and reduced generalization to unseen temperature conditions, as the model fails to capture the nuanced and specific effects of temperature on different types of sensor data.

\subsection{Ablation Study}
\begin{figure}[tbh]
  \centering
    \includegraphics[width=.8\linewidth]{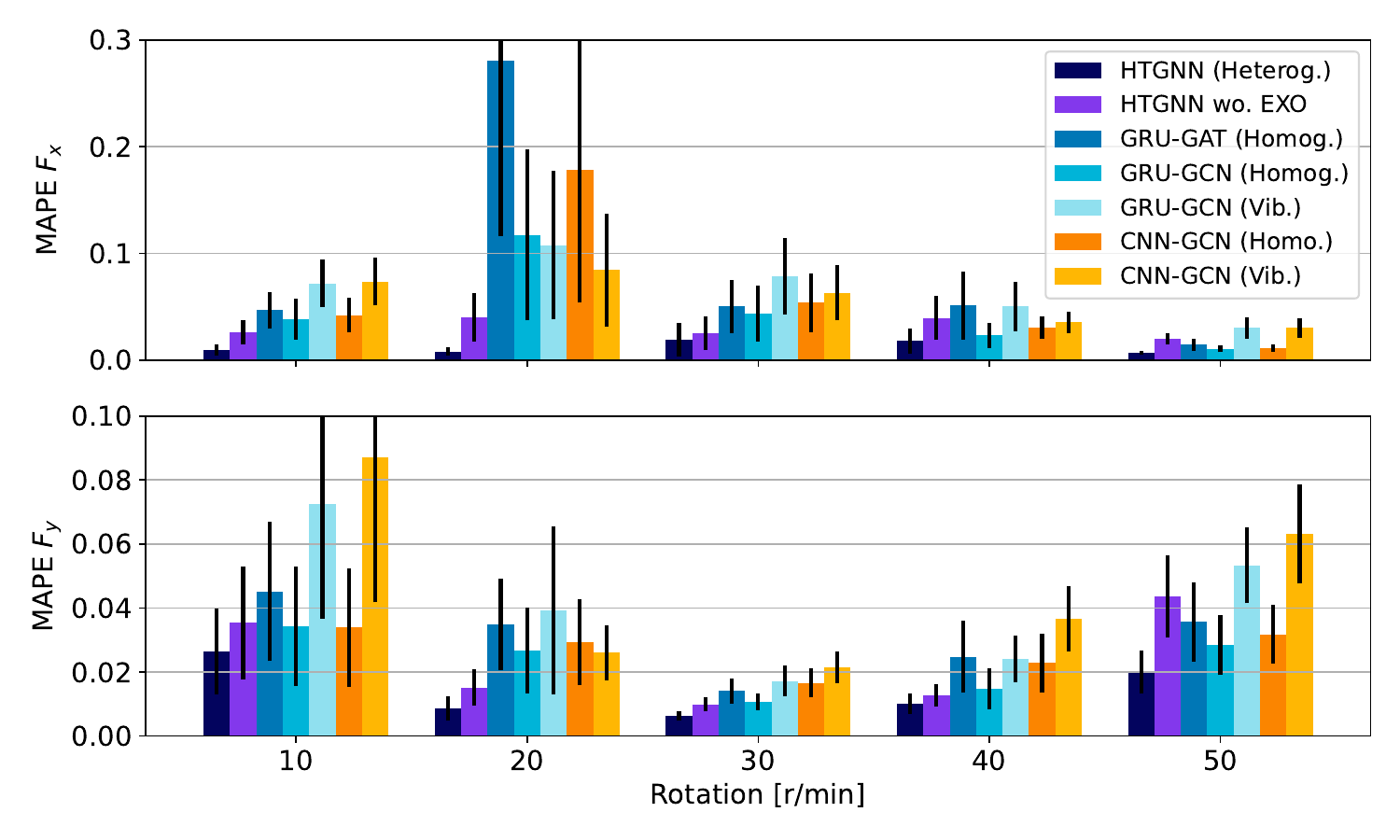}
  \caption{Ablation study on the bearing dataset with five runs}
  \label{fig:ablation}
\end{figure}
To assess the individual contributions of HTGNN components and the impact of different input modalities, we conducted an ablation study on the bearing dataset. We compared HTGNN against several simplified models, evaluating their performance using MAPE as the metric. The configurations tested include:
\begin{enumerate}[itemsep=5pt,topsep=5pt,parsep=0pt,partopsep=0pt]
\item \textbf{HTGNN without Exogenous Variables (HTGNN wo. EXO):} This configuration removes the explicit operating condition modeling in the dynamics extraction, incorporating the embeddings of exogenous variables directly into the graph head.
\item \textbf{Homogenous graph configurations:} These models use a homogenous graph where all nodes (temperature, vibration, and speed) are treated equally and edge types are not differentiated. 
   \begin{itemize}[itemsep=5pt,topsep=5pt,parsep=0pt,partopsep=0pt]
   \item \textbf{GRU-GAT (Homog.):} Leverages GRU for extracting temporal dynamics, particularly effective for low-frequency signals, and GAT to weigh the importance of nodes in the aggregation process.
   \item \textbf{GRU-GCN (Homog.):}: Employs the same GRU encoder as GRU-GAT for capturing temporal dynamics but uses GCN for aggregation, assuming equal importance across nodes.
   \item \textbf{CNN-GCN (Homog.):} Employs 1D-GCNN, incorporating a gating mechanism to modulate the extraction of frequency-domain features from vibration signals, followed by GCN aggregation, assuming equal feature importance.
   \end{itemize}
\item \textbf{Vibration-only configurations:}
   \begin{itemize}[itemsep=5pt,topsep=5pt,parsep=0pt,partopsep=0pt]
   \item \textbf{CNN-GCN (Vib.):} Identical to HTGNN but removes temperature nodes and heterogeneous interaction modeling.
   \item \textbf{GRU-GCN (Vib.):} Similar to CNN-GCN (Vib.) but uses a GRU encoder instead of 1D-GCNN.
   \end{itemize}
\end{enumerate}
We also experimented with temperature-only configurations using a GRU-GCN architecture. However, these configurations performed significantly worse than the other models, highlighting the critical importance of vibration signals for accurate load estimation. Due to their poor performance, the results for these temperature-only models are not included in the subsequent analysis.
Fig.~\ref{fig:ablation} presents the results of the ablation study, demonstrating the impact of removing components or input modalities on the performance of the HTGNN model. As depicted in the figure, the removal of temperature information (comparing the vibration-only configurations to HTGNN) significantly degrades the accuracy of radial load predictions, particularly at extreme rotational speeds. This observation aligns with the insights from Fig.~\ref{fig:temp_diff_distribution}, which indicates that temperature change rates are more indicative of radial load distribution than axial load.
 
\textbf{Impact of exogenous variable modeling.} 
Furthermore, the omission of explicit exogenous variable modeling (HTGNN wo. EXO) leads to a noticeable decrease in performance across all rotational speeds for both axial and radial load predictions. This finding underscores the importance of incorporating operating conditions as contextual information to capture the complex dynamics of the bearing's behavior under diverse loads and speeds. Incorporating these variables helps the model understand and adapt to varying operating conditions, thereby enhancing prediction accuracy and robustness.

\textbf{Homogenous vs. heterogeneous interaction modeling.}
The performance difference between the homogeneous graph configurations (GRU-GAT, GRU-GCN, CNN-GCN) and HTGNN (Heterog.) emphasizes the benefits of employing a heterogeneous graph structure. Interestingly, GRU-GAT, which uses an attention mechanism to assign weights to different nodes, does not necessarily outperform the other homogeneous models (GRU-GCN, CNN-GCN). In our experiments, GRU-GAT consistently performed worse than GRU-GCN. This suggests that simply weighting nodes differently, as done by GAT, might not be as effective as explicitly distinguishing between different types of nodes (temperature, vibration) and edges (interaction types) within the graph structure. The heterogeneous approach enables HTGNN to learn more meaningful and informative representations of the bearing system, ultimately leading to improved prediction accuracy.

\textbf{Vibration vs. temperature.} 
Both vibration-only configurations exhibit reduced accuracy for load prediction compared to their homogeneous counterparts, especially at extreme rotational speeds. This observation underscores the importance of temperature information, which provides valuable insights into load distribution in the bearing, especially for the radial loads that are not fully captured by vibration signals alone.  
Among the vibration-only models, GRU-GCN (Vib.) shows superior radial load prediction, likely due to its ability to capture vibration magnitude trends over time. Conversely, CNN-GCN (Vib.) demonstrates an advantage in axial load prediction, possibly due to its focus on frequency-domain features.

Overall, this ablation study highlights the importance of key design choices in HTGNN: explicit modeling of exogenous variables (operating conditions), and differentiated node and edge types for effective fusion of heterogeneous sensor information. These elements are crucial for maximizing performance in bearing load estimation and offer valuable insights for the development of effective virtual sensors in industrial applications with mixed modalities.

\section{Conclusion}
\label{sec:conclusion}
In this research, we have presented HTGNN, a novel GNN-based virtual sensor framework specifically designed to address the challenges of heterogeneous temporal dynamics and varying operating conditions in complex systems.
Through extensive experiments on both a test-rig bearing dataset, characterized by high-frequency vibrations and temperature fluctuations under varying rotational speeds, and a simulated train-bridge-track interaction dataset, capturing multi-scale displacement and acceleration data influenced by ambient temperature variations, we have demonstrated HTGNN's superior performance and adaptability in virtual load sensing across diverse industrial domains, highlighting its potential for broad applicability.
HTGNN's effectiveness stems from its ability to explicitly model the complex, heterogeneous relationships between sensor modalities and its capacity to extract operating condition-aware dynamics. This approach allows HTGNN to adapt to changing operating conditions and accurately predict loads even in scenarios where traditional methods struggle, such as low rotational speeds for bearings or extreme temperatures for bridges. The consistent robustness and accuracy of HTGNN in both case studies highlight its potential as a reliable virtual sensor for diverse IIoT applications, enabling effective monitoring, predictive maintenance, and enhanced system performance.
Our findings highlight the importance of leveraging physical prior and domain knowledge in the design of virtual sensors, particularly through the use of GNNs. By encoding domain knowledge and physical priors about system behavior into the construction of the graph structure, GNNs can explicitly model the physical relationships between sensors, providing the model with valuable insights into load distribution across the system and leading to significantly improved prediction accuracy.
\revtext{Furthermore, HTGNN demonstrates promising computational efficiency and scalability. By sharing encoders for all nodes of the same type, the model can compute node embeddings in parallel. This parallel processing, combined with the use of efficient GNN operations, significantly reduces the overall computation time. Consequently, HTGNN can efficiently handle large-scale industrial systems with many sensors. The model's ability to perform fast inference enables real-time load prediction, which is crucial for applications such as real-time monitoring and control in industrial settings.}
While this work focuses on virtual sensing for load prediction, the underlying principles of HTGNN, such as heterogeneous interaction modeling and operating condition-aware dynamics extraction, are generalizable and can be applied to other downstream tasks within the IIoT domain. For example, HTGNN could be adapted for tasks like signal reconstruction, anomaly detection, or forecasting, where the ability to handle heterogeneous temporal dynamics and varying operating conditions is crucial.

\section{Data and Code Availability}
The bearing test rig dataset is available at \url{https://doi.org/10.5281/zenodo.14959001}, and the bridge dataset can be accessed at \url{https://doi.org/10.5281/zenodo.14972955}. The code repository at \url{https://github.com/EPFL-IMOS/htgnn} contains data loaders for both datasets, along with the implementation of the proposed method and the baseline methods.

\section*{Acknowledgments}
This work was supported by the Swiss National Science Foundation under Grant 200021\_200461. The authors would like to thank Muhammad Zohaib Sarwar for his valuable insights and discussions regarding the TTB simulator setup. Additionally, the authors gratefully acknowledge the use of AI-powered tools for refining the text of this manuscript.


\appendix
\section{Details of Case Studies and Evaluation Setups}

\subsection{Bearing Operating Conditions}
\label{sec:app_bearing_conds}
\begin{figure}[tbhp]
  \centering
    \includegraphics[width=.75\linewidth]{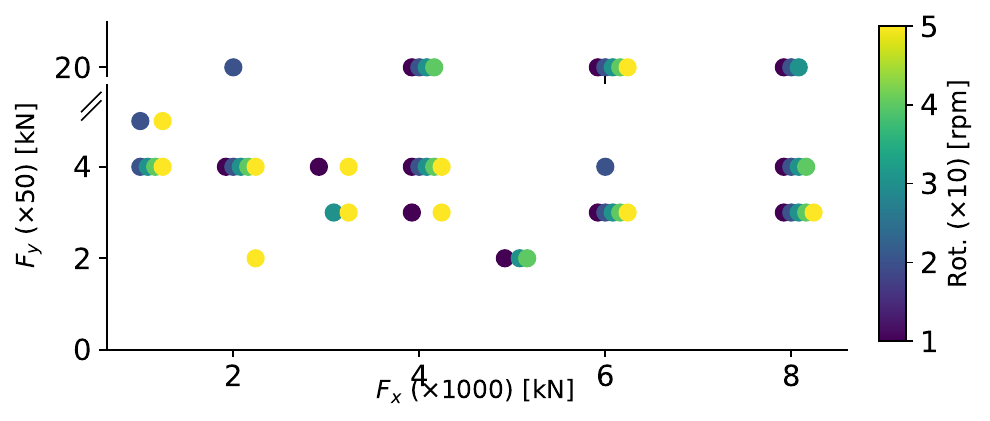}
  \caption{Operating conditions considered in the bearing load prediction case study.}
  \label{fig:bearing_cond}
\end{figure}
Fig.~\ref{fig:bearing_cond} illustrates the 55 unique operating conditions under which the bearings were tested, defined by axial load ($F_x$), radial load ($F_y$), and rotational speed.

\subsection{Heterogeneous Bridge Graph Construction}
\label{sec:app_bridge_setup}
\begin{figure}[tbhp]
  \centering
    \includegraphics[width=.75\linewidth]{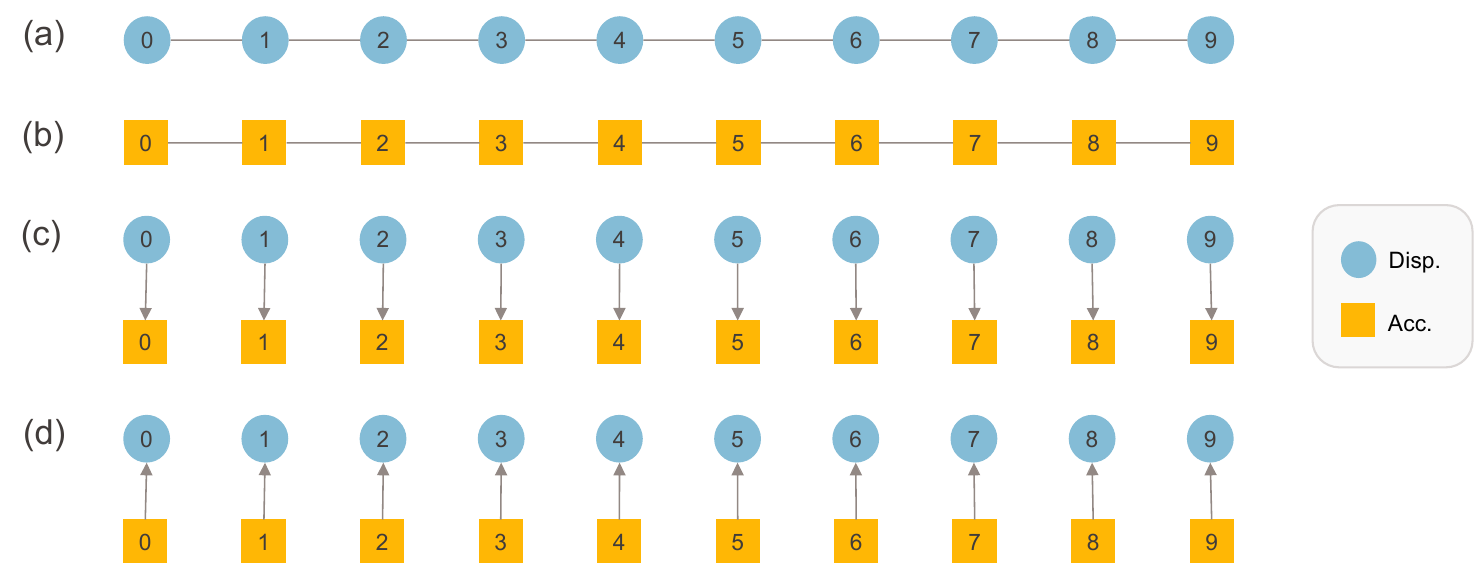}
  \caption{Heterogeneous graphs for bridge sensor network relationship modeling. (a) D-D (b) A-A (c) D-A (d) A-D}
  \label{fig:bridge_graphs}
\end{figure}
We construct a heterogeneous graph for the bridge sensor network, following the same principles established for the bearing graph. Nodes in the graph represent two types of sensors: displacement (D) and acceleration (A).
The node positions within the graph reflect the physical placement of sensors on the bridge structure. 
Four types of relationships are modeled: D-D, A-A, D-A, and A-D. Homogeneous relationships (D-D and A-A) connect sensors of the same type based on their direct spatial proximity. 
Heterogeneous relationships (D-A and A-D) are also established based on direct proximity, connecting displacement, and acceleration sensors installed at the same location.
Figures \ref{fig:bridge_graphs}(a) and \ref{fig:bridge_graphs}(b) illustrate the connectivity within a single sensor type based on spatial proximity.  We assume symmetrical (undirected) relationships for homogeneous connections. Heterogeneous relationships (D-A and A-D) are modeled with directed edges, as shown in Figures \ref{fig:bridge_graphs}(c) and \ref{fig:bridge_graphs}(d), to capture the potential influence of displacement on acceleration measurements.

\subsection{Hyperparameter Tuning}
\label{sec:app_hyper}

\begin{table*}[tbhp]
\centering
\caption{Range of hyperparameters with optimal values indicated within parentheses.}
\renewcommand{\arraystretch}{1.1} 
\resizebox{1\columnwidth}{!}{%
    \label{table:hyperparameters}
    \begin{tabularx}{1.62\linewidth}{l *{6}{c}} 
    \toprule
    \textbf{Model} & \textbf{L} & \textbf{H} & \textbf{Norm.} &\textbf{$p$} & \textbf{Additional Parameters} & \textbf{{No. Params}} \\ \midrule
    BiLSTM & 1-2 ($1$) & 10-50 ($50$) & / & 0-0.5 (0.2)& - & $41852 \mid 39401$ \\
    1D-CNN & 3-5 ($4$) & 20-50 ($50$) & BN & 0-0.5 (0.2) & kernel 3-9 ($9$) & $59703 \mid 63552 $ \\
    1D-GCNN & 2-$4^*$ & $20-50^*$ & BN & 0-0.5 (0.2) & dil. 1-2, cont. dim. 1-10 ($5\mid 10$) & $39620 \mid 39169$ \\
    \revtext{TimeMixer} & \revtext{2-6 ($2\mid4$)} & \revtext{16-64 ($16\mid32$)} & \revtext{/$^*$} & \revtext{0-0.5 (0.1)} & \revtext{decomp./downs. method$^*$, downs l. 1-3 ($2$)} &  \revtext{$12416 \mid 41877$} \\
    MTGAT & 1-2 (1) & 50-100 ($50$) & BN & 0-0.5 ($0.5\mid 0$) & kernel 5-9 (7), att. emb. 50-100 (100) & $65711 \mid 66182$ \\
    HTGNN & 3 & 10-20 ($10\mid20$) & / & 0-0.5 (0.2) & gnn emb. 20-40 ($40$) & $34860 \mid 40365 $ \\
    \bottomrule
    \end{tabularx}
}
\end{table*}
Initial hyperparameters\revtext{, including the selection of data preprocessing parameters (window size and stride),} were drawn from relevant literature and subsequently refined through a grid search over an expanded parameter space tailored to the specific sensitivities of our case studies. 
\revtext{\textbf{Data preprocessing parameters}:
For each dataset, the window size and stride used for data preprocessing were empirically determined. To ensure a fair comparison across all methods, these values were determined based on preliminary experiments with the baseline 1D-CNN model and then kept consistent for all other baseline models as well as our proposed method. For the window size, we tested values of 30 and 60. The optimal size varied between case studies, likely due to differences in sampling frequency and the nature of the underlying dynamic. Specifically, a window size of 60 was optimal for the bearing case study, while 30 was optimal for the train-track-bridge case study. The stride was also adjusted per dataset. We used a stride of 1 for the bearing case study and 5 for the train-track-bridge case study, reflecting the larger size of the latter. 
\textbf{Model hyperparameters:}}
Table~\ref{table:hyperparameters} summarizes the optimal \revtext{model} configurations for each case study.
Values in parentheses indicate the best configuration for the respective case study; a single value denotes the optimal choice for both. 
An asterisk ($^*$) indicates additional details provided in the text.
The grid search explored variations in the number of layers (L), hidden layer dimensions (H), normalization techniques (Batch Normalization (BN) and Layer Normalization (LN)), and dropout rates $p$ (ranging from 0 to 0.5). The hyperparameter search space was designed to maintain a roughly consistent number of parameters across all models. Model-specific parameters and optimal model sizes are also detailed in Table~\ref{table:hyperparameters}. The final hyperparameter selections were based on achieving the lowest validation loss. 
Specifically, the hyperparameter tuning process for each baseline model is as follows:
\begin{itemize}[itemsep=5pt,topsep=5pt,parsep=0pt,partopsep=0pt]
    \item \textbf{BiLSTM}:
    The number of LSTM layers and hidden layer dimensions were varied. Layer normalization was explored but found to not improve performance.
    \item \textbf{1D-CNN}:
    Adapted from the design in \cite{chao2022fusing}, this model's hyperparameter search focused on hidden channel dimension, kernel size, number of channels, and number of layers.
    \item \textbf{G-CNN}:
     A gated version of 1DCNN, similar to HTGNN, this model also consists of separate high- and low-frequency convolutional channels. In each channel, convolutional layer outputs are element-wise multiplied by a gating signal derived from the exogenous variable embedding. The optimal configuration utilized a kernel size of 3 and a dilation of 1 for high-frequency channels, and a kernel size of 5 and dilation of 2 for low-frequency channels. The hyperparameter search explored varying channel numbers, with the optimal configuration identified as 50-50-1 for both case studies. Additionally, different exogenous variable embedding sizes were explored.
     \item \revtext{\textbf{TimeMixer}: Adapted from \cite{wang2024deep}, we varied encoder layers (2-6). The hidden dimension was varied from 16-64, with the final fully connected layer's dimension set as double. Averaging was the best downsampling strategy for the bearing case study and convolution for the bridge case study. For both, 2 downsampling layers (1-3 tested) were optimal.  Moving average decomposition was best for the bearing dataset, while Discrete Fourier Transform (DFT) decomposition with 9 top frequencies (3-15 tested) was best for the bridge dataset. Reversible Instance Normalization (RevIN)~\cite{kim2021reversible} normalization was not found to be useful.
    }
    \item \textbf{MTGAT}:
    The attention module was upgraded to GATv2 \cite{brody2022gatv2} for enhanced expressivity. The hyperparameter search explored its 1DCNN encoder kernel sizes ranging from 5 to 9, as well as hidden dimensions for feature and temporal attention embeddings between 50 and 100. Additionally, the number of GRU layers (1 or 2) and the GRU hidden dimension (50 to 100) were tested. The original reconstruction and forecasting modules were removed and replaced with an MLP head for a direct graph-level output.
    \item \textbf{HTGNN}:
    To reduce the search space, we maintained a consistent hidden size across all layers and the same graph embedding dimension for all GNN modules. 
    The search focused on varying the hidden node dimension (10 to 20), the hidden graph dimension (20 to 40), and the head hidden dimension (20 to 40). For the high-frequency node encoder, we adopted the same kernel size and dilation configuration as G-CNN, but with channel sizes of 4-4-1. The number of heterogeneous GNN layers was fixed at 3.
\end{itemize}

\bibliographystyle{IEEEtran}
\bibliography{references}

\end{document}